\documentclass[10pt,twocolumn,letterpaper]{article}

\usepackage{amsmath,amssymb,amsbsy,amsfonts,dsfont,pifont,bm,bbm,mathrsfs,mathtools,nicefrac}
\usepackage{algorithm,algpseudocode,listings}
\usepackage{multirow,adjustbox,diagbox,threeparttable,tabularray}
\usepackage{wacv}              %
\usepackage[accsupp]{axessibility}  %

\usepackage[table]{xcolor}
\definecolor{citeblue}{rgb}{0.21,0.49,0.74}

\usepackage{wrapfig}

\definecolor{wacvblue}{rgb}{0.21,0.49,0.74}
\definecolor{darkgreen}{rgb}{0.0,0.5,0.0}
\definecolor{lightgreen}{HTML}{E6F4EA}
\usepackage[pagebackref,breaklinks,colorlinks,allcolors=wacvblue]{hyperref}
\usepackage[capitalize]{cleveref}
\def\wacvPaperID{2259} %
\def\confName{WACV}
\def\confYear{2027}
\newcommand{\modelname}{Mem3R\xspace}
\newcommand{\ours}{Ours\xspace}%

\newcommand{\gb}{\cellcolor{lightgreen}}
\title{\modelname: Streaming 3D Reconstruction with Hybrid Memory}

\author{%
  Changkun Liu$^{1,2,*,\dagger}$
  \quad
  Jiezhi Yang$^{1,*}$
  \quad
  Zeman Li$^{1}$
  \quad
  Yuan Deng$^{1}$
  \\
  Jiancong Guo$^{1}$
  \quad
  Binbin Xu$^{1}$
  \quad
  Luca Ballan$^{1}$\\
  $^1$Google \quad
  $^2$The Hong Kong University of Science and Technology 
}

\begin{document}

\twocolumn[{

\maketitle
\centering
\includegraphics[width=0.85\linewidth]{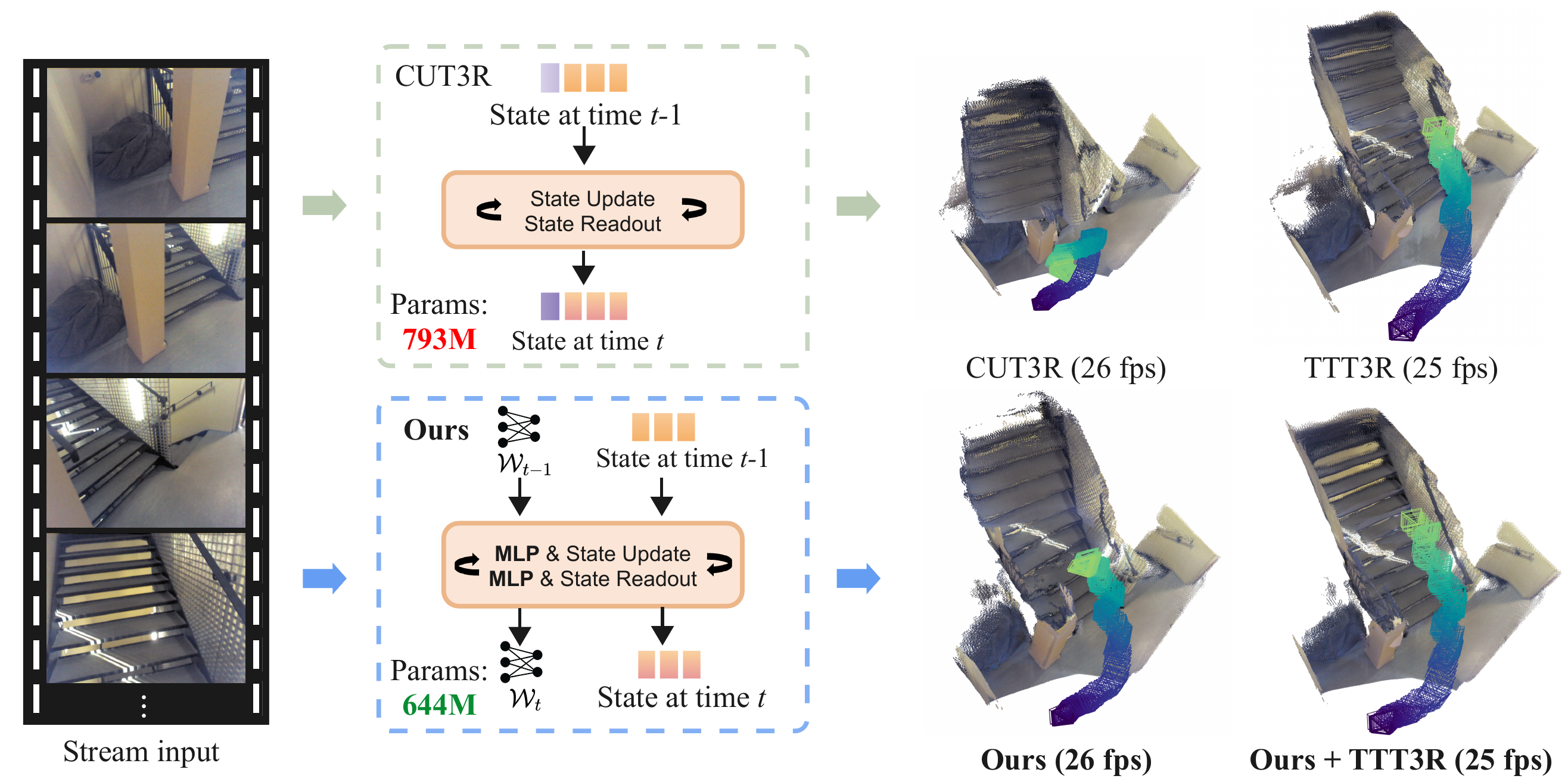}
\captionsetup{type=figure}
\captionof{figure}{
\textbf{\modelname} is an RNN-based model built on the CUT3R paradigm
that achieves stronger long-sequence streaming 3D perception.
Built on a dual-memory design, \modelname combines
(i) an \textit{implicit memory} $\mathcal{W}$ for camera pose estimation and
(ii) an \textit{explicit memory} of persistent tokens for global geometric context.
\modelname is further compatible with plug-and-play state-update strategies
developed for CUT3R, such as TTT3R.
}
\label{fig:teaser}
\par\bigskip
}]

\begingroup
\renewcommand{\thefootnote}{\fnsymbol{footnote}}
\footnotetext[1]{Equal Contribution.}
\footnotetext[2]{Work done during a student researcher internship at Google.}
\endgroup

\begin{abstract}
Streaming 3D perception is well suited to robotics and augmented reality, where long visual streams must be processed efficiently and consistently. Recent recurrent models offer a promising solution by maintaining fixed-size states and enabling linear-time inference, but they often suffer from temporal forgetting over long sequences due to the limited capacity of compressed memories. 
We propose \textbf{\modelname}, a streaming 3D reconstruction model with a hybrid memory design that decouples camera tracking from geometric mapping to improve temporal consistency over long sequences. For camera tracking, \modelname employs an implicit fast-weight memory implemented as a lightweight Multi-Layer Perceptron updated via Test-Time Training. For geometric mapping, \modelname maintains an explicit token-based fixed-size state. Compared with CUT3R, this design not only significantly improves long-sequence performance but also reduces the model size from 793M to 644M parameters. \modelname supports existing improved plug-and-play state update strategies developed for CUT3R. Specifically, Mem3R combined with TTT3R reduces absolute trajectory error by up to 39\% relative to CUT3R combined with TTT3R on 500- to 1000-frame sequences from the TUM Dynamics dataset. The resulting improvements also extend to other downstream tasks, including video depth estimation and 3D reconstruction. Project page: \url{https://lck666666.github.io/Mem3R/}.
\end{abstract}

\section{Introduction}
\label{sec:intro}
A central challenge in real-time 3D perception is to recover accurate scene geometry and camera motion from long visual streams under strict computational and memory budgets. This capability is essential for a wide range of mobile applications, including robotic visual localization, navigation, and augmented reality. Recent progress in feed-forward multi-view geometry, driven by models such as DUSt3R~\cite{DUSt3R} and VGGT~\cite{wang2025vggt}, has inspired a new wave of Transformer-based architectures~\cite{leroy2024grounding,tang2025mv,yang2025fast3r,zhang2025flare,wang2026pi,shen2025fastvggt} that achieve impressive performance in 3D reconstruction and pose estimation. However, these methods typically rely on global pairwise alignment or dense cross-frame attention, leading to computational costs that scale quadratically with sequence length and limit their applicability in long-horizon streaming scenarios.

To enable scalable streaming, prior work has mainly evolved along two directions. The first  line of work leverages causal-attention Key-Value (KV) caches to preserve the full observation history~\cite{wang20253d,lan2026streamr,yuan2026infinitevggt,zhuo2025streaming}. While effective, this paradigm incurs memory consumption that grows linearly with sequence length, making it increasingly vulnerable to out-of-memory (OOM) failures during long-horizon inference. The second direction, exemplified by CUT3R, adopts a Recurrent Neural Network (RNN)-based architecture with a fixed-size latent state, enabling constant-memory inference. However, this compact state also creates an information bottleneck: once the sequence length exceeds the typical training horizon, CUT3R suffers from substantial temporal forgetting and trajectory drift. To alleviate these issues, TTT3R~\cite{chen2026tttr} reinterprets state updates through the perspective of Test-Time Training (TTT) and introduces per-token update weights that emphasize tokens most relevant to the current observations. Building on this idea, subsequent works~\cite{zheng2026ttsa3r,dong2026memix} have proposed alternative plug-and-play update strategies to further improve CUT3R. 

In this work, we present \textbf{\modelname}, an enhanced RNN-based architecture with a dual-memory design for long-horizon streaming 3D perception. Our key idea is to improve the representational efficiency of the recurrent state by decoupling camera tracking from geometric mapping. Specifically, \modelname replaces CUT3R's pose-related token memory and its Transformer-based memory read/write blocks (implemented as the \texttt{LocalMemory} module in the official CUT3R codebase) with a lightweight implicit Multi-Layer Perceptron (MLP)-based memory for pose estimation via TTT inspired by LaCT~\cite{zhang2025test}, while retaining an explicit token-based memory to preserve global geometric context. This dual-memory design yields a more compact and effective backbone, reducing the parameter count from 793M to 644M (a 19\% reduction) while achieving stronger performance, as illustrated in \cref{fig:teaser}. We also introduce a learnable channel-wise state update module, inspired by recent advances in finite-state RNN architectures~\cite{team2025kimi,behrouz2024titans}, to enable finer-grained control over state updates.

As a result, \modelname remains fully compatible with training-free per-token state-update strategies such as TTT3R~\cite{chen2026tttr} and TTSA3R~\cite{zheng2026ttsa3r}, and it delivers further gains when combined with them. Across camera pose estimation, video depth estimation, and 3D reconstruction, \modelname shows clear improvements in most evaluated settings while preserving the inference speed and constant-memory advantages of RNN-based streaming models. These results establish \modelname as a stronger and more memory-efficient backbone for long-sequence streaming 3D perception.

\section{Related Work}
\begin{figure*}[htbp]
\centering
\includegraphics[width=0.95\textwidth]{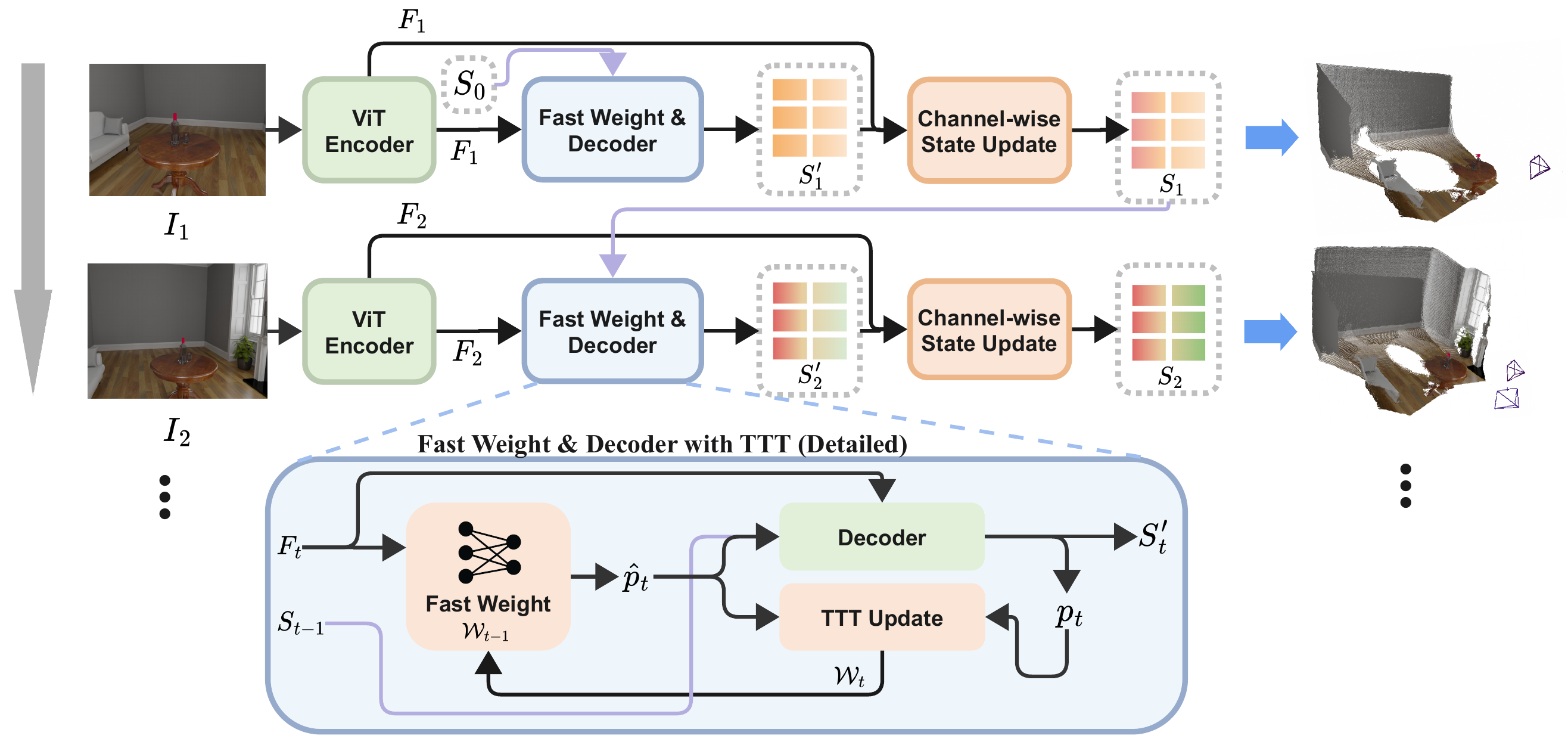}
\caption{
Overview of \modelname. \textbf{Top:} For each frame $I_t$ in the streaming image sequence, a ViT encoder extracts image features $F_t$. The MLP-based fast-weight module $\mathcal{W}$ performs camera tracking with decoder under Test-Time Training (TTT), while the fixed-size state $\mathcal{S}$ and decoder preserve and update geometric information, producing an intermediate state $S_t'$. $S_t'$ is then fused with the previous state $S_{t-1}$ through a channel-wise update gate module (see \cref{subsec:channel_wise}) to obtain the final updated state $S_t$ . By replacing pose-related state token with $\mathcal{W}$, \modelname maintains a constant-size recurrent state while reducing long-term temporal forgetting in streaming 3D reconstruction. \textbf{Bottom:} Illustration of the fast-weight and decoder module with TTT (see \cref{subsec:fast_weight}).}

\label{fig:framework}
\end{figure*}
\paragraph{Offline Feed-forward 3D Reconstruction.}
DUSt3R~\cite{DUSt3R} proposes an end-to-end, feed-forward approach to 3D reconstruction, directly regressing point maps from image pairs. Subsequent extensions~\cite{leroy2024grounding,jayantisegmast3r,liuplana3r,fan2024large,chen2025easi3r,zhang2025monstr,xusiu3r, chen2024pref3r,fang2026densr} broadened this formulation to more challenging settings, but most remain fundamentally pairwise and therefore require costly global alignment once more than two views are involved. To overcome this limitation, recent offline multiview models~\cite{zhang2025flare,yang2025fast3r,wang2025vggt,wang2026pi,shen2025fastvggt,gao2025more,keetha2025mapanything} employ large feed-forward Transformers with global attention to jointly process all input views and predict per-view geometry and camera poses in a single forward pass. Representative examples such as VGGT~\cite{wang2025vggt} achieve strong reconstruction quality, but their computational and memory costs scale quadratically with the number of frames. Moreover, they are inefficient for streaming data,  as incorporating a newly arrived view typically requires reprocessing the entire sequence. Although methods such as FastVGGT~\cite{shen2025fastvggt} reduce inference overhead through token merging, these approaches remain better suited to offline reconstruction than to real-time long-sequence streaming settings.

\paragraph{Online 3D Reconstruction.}
Online 3D reconstruction methods~\cite{wang20253d,yuan2026infinitevggt,lan2026streamr,zhuo2025streaming,wang2025continuous,wu2025point3r,li2026wintr,chen2025long3r} process views incrementally and update scene representations on the fly, making them attractive for robotics and embodied perception. Existing approaches mainly differ in how they preserve historical context. Causal-cache methods, such as StreamVGGT~\cite{zhuo2025streaming}, retain past keys and values for long-range temporal conditioning, but their memory and computation still grow with sequence length. In contrast, recurrent methods such as CUT3R~\cite{wang2025continuous} maintain a fixed-size latent state for constant-memory inference. However, compressing long-term observations into a compact state creates an information bottleneck, often causing temporal forgetting and drift on long sequences. Recent training-free update strategies, including TTT3R~\cite{chen2026tttr} and TTSA3R~\cite{zheng2026ttsa3r}, improve CUT3R-style recurrent updates, while Point3R~\cite{wu2025point3r} enhances memory recall through explicit 3D point associations at the cost of memory growth. In contrast, our method improves the representational efficiency of the recurrent backbone through a dual-memory design, while remaining compatible with plug-and-play update strategies such as TTT3R and TTSA3R.

\paragraph{Fast Weight Programs and Memory Architectures.}
The view of linear layers as associative memories dates back to Hopfield networks~\citep{hopfield1982neural} and was later developed in fast weight programmers, where dynamic fast programs are integrated into recurrent neural networks as writable memory stores~\citep{schlag2021linear, schmidhuber1992learning, schmidhuber1993reducing}. Building upon these foundational concepts, modern sequence modeling has diverged into two primary structural paradigms: \emph{linear memory modules} and \emph{deep memory modules}. \emph{Linear memory modules}~\citep{gu2024mamba, peng2023rwkv, yang2024parallelizing, yang2024gated, hu2025comba, team2025kimi} maintain matrix-valued hidden states with linear state transitions, while \emph{deep memory modules}~\citep{behrouz2024titans, behrouz2025nested, li2025tnt, sun2024learning, behrouz2026its, behrouz2025atlas, zhang2025test} use online-adapted sub-networks with dynamically updated parameters to encode contextual information. Our method is particularly inspired by this latter line of work, using an online-adapted implicit memory to improve long-horizon performance. Concurrent works~\cite{wang2026tttlrm,zhang2026loger,jin2026zipmap} also incorporate TTT memory layers into 3D perception but address different settings: tttLRM~\cite{wang2026tttlrm} targets novel view synthesis with 3D Gaussian Splatting, ~\cite{jin2026zipmap,elflein2026vggttt,Xie_2026_CVPR} focus on faster offline reconstruction, and LoGeR~\cite{zhang2026loger} studies chunk-wise reconstruction with sliding-window attention rather than frame-by-frame streaming reconstruction.

\section{Method}
In streaming 3D perception, a model processes an image sequence $\{I_t\}_{t=1}^{N}$ incrementally, estimating camera motion and scene geometry online as new observations arrive.
\subsection{Overview}
\label{subsec:overview}

We begin by briefly reviewing the recurrent state formulation in CUT3R to highlight the key differences in our design:
\begin{equation}
    X'_t, \bar{S}_t = \mathrm{Decoders}(X_t, S_{t-1}),
    \label{eq:cut3r_decoder}
\end{equation}
where $S_{t-1}$ is the state at time $t-1$, $\bar{S}_t$ is the updated state produced directly by the decoder, and $X_t=[z_t, F_t]$ consists of a learnable pose token $z_t$ and image tokens $F_t$ extracted from the current frame $I_t$.

 Our \modelname is also an RNN-based architecture that maintains a fixed-size recurrent state for online 3D perception. The key idea is to improve the efficiency of the recurrent memory representation by replacing the pose-related components of the recurrent state and transformer-based decoder blocks with an implicit MLP. To this end, \modelname uses a \emph{hybrid memory} design: an \emph{implicit memory} $\mathcal{W}$, implemented as an MLP-based fast-weight module for camera tracking, and an \emph{explicit memory} $S$, implemented as persistent state tokens for geometric mapping, as shown in \cref{fig:teaser} and \cref{fig:framework}. Unlike CUT3R, which directly writes decoder outputs into the recurrent state, our model treats the state output from the decoder as a candidate update and integrates it with a learnable channel-wise gate.

At time $t$, the current frame is encoded into visual tokens $F_t$. The implicit memory predicts a pose token prior $\hat{p}_t$, which is processed together with $F_t$ and the previous explicit state $S_{t-1}$ by a transformer decoder. The decoder produces a posterior pose token $p_t$, updated visual features $F'_t$, and a candidate state update $S'_t$. The implicit memory is then updated online by aligning prior $\hat{p}_t$ with posterior $p_t$, while the explicit memory is updated through channel-wise gated integration of $S'_t$. 
\subsection{Implicit Memory via Test-Time Training}
\label{subsec:fast_weight}

To improve long-horizon camera tracking while preserving a constant inference speed, we introduce an adaptive fast-weight memory implemented as a lightweight SwiGLU MLP~\cite{shazeer2020glu}:
\begin{equation}
    f_{\mathcal{W}}(x) = W_2 \Big( \mathrm{SiLU}(W_1 x) \odot (W_3 x) \Big),
\end{equation}
where $\mathcal{W}=\{W_1,W_2,W_3\}$ denotes the fast weights updated online during inference.

Given the current visual tokens $F_t$, we first compute a $L_2$-normalized query feature $\kappa_t$ and read a pose prior from the previous fast weights $\mathcal{W}_{t-1}$:
\begin{equation}
    \hat{p}_t = f_{\mathcal{W}_{t-1}}(\kappa_t).
    \label{eq:hat_p_t}
\end{equation}
This prior replaces CUT3R's learned pose token $z_t$, yielding $
    X_t = [\hat{p}_t, F_t].
$
The decoder then produces
\begin{equation}
    X'_t, S'_t = \mathrm{Decoders}(X_t, S_{t-1}),
    \label{eq:decoder}
\end{equation}
where $X'_t=[p_t, F'_t]$, with $p_t$ defined as the refined posterior pose token and $F'_t$ as the state-enriched visual features.

We update the implicit memory by aligning the prior and posterior:
\begin{equation}
    \mathcal{L}_{\mathrm{ttt}}(\hat{p}_t, p_t) = \langle \hat{p}_t, p_t \rangle.
\end{equation}
To make the update adaptive for different $I_t$, we predict a per-step decay factor $\alpha_t = 1 - \gamma \cdot \sigma(W_{\alpha} F_t)$ and per-layer learning rates $\eta_t$ from the current features:
\begin{equation}
    \eta_t = \mathrm{Softplus}\!\left(
        W_{\eta} F_t + \mathrm{Softplus}^{-1}(\eta_{\mathrm{base}})
    \right),
\end{equation}
where $\gamma=0.01$ is the decay scale and
$\eta_{\mathrm{base}}=0.001$ is the base learning rate.
Here, $\mathrm{Softplus}^{-1}(x)=\log(e^x-1)$ ensures that
$\eta_t=\eta_{\mathrm{base}}$ when $W_{\eta}F_t=0$. $\sigma(\cdot)$ denotes the sigmoid function. The fast weights are then updated by
\begin{equation}
    \mathcal{W}_t
    =
    \alpha_t \mathcal{W}_{t-1}
    +
    \eta_t \odot \nabla_{\mathcal{W}} \mathcal{L}_{\mathrm{ttt}}.
    \label{eq:fast_weight_update}
\end{equation}
This update enables the model to estimate accurate pose cues while filtering out transient noise, effectively turning the fast-weight module into a compact implicit temporal memory for robust camera tracking. We provide more details about this module in the supplementary material.

\begin{figure*}[h!]
\centering
\includegraphics[width=0.9\textwidth]{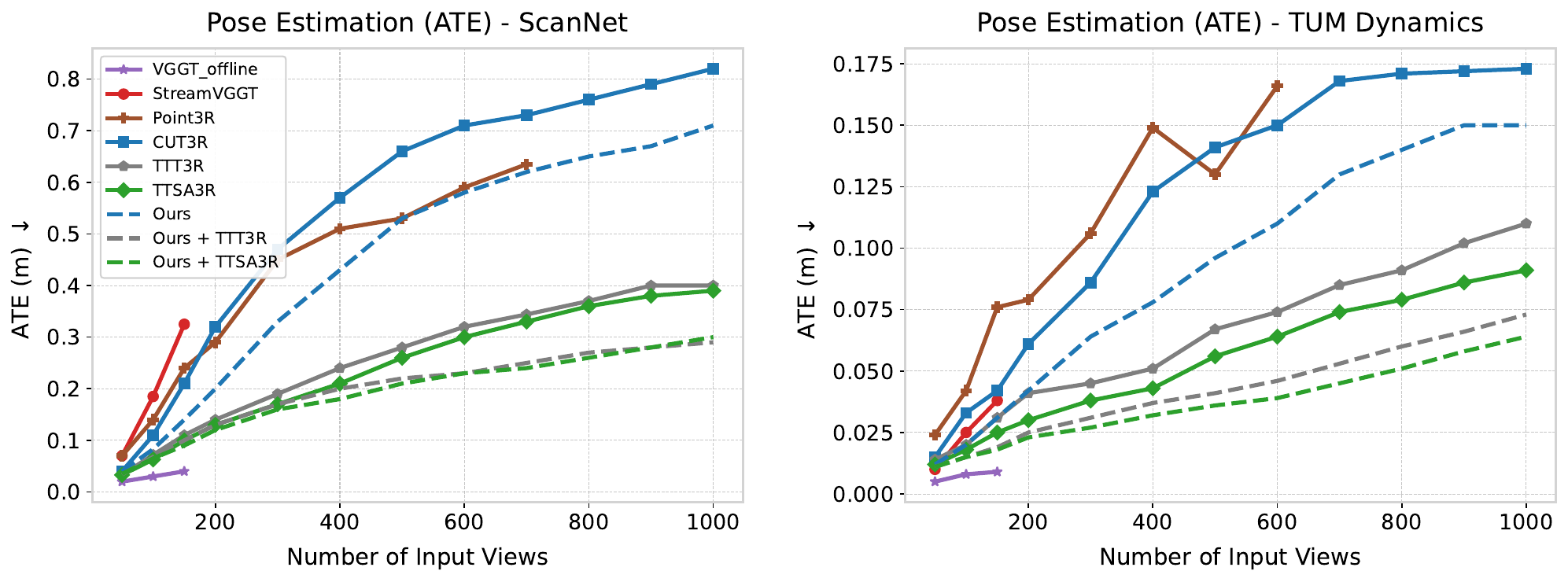}
\caption{Quantitative evaluation of camera pose estimation from the ScanNet dataset (left) and the TUM Dynamics dataset (right). \modelname achieves significant improvement compared with CUT3R. Moreover, when combined with the training-free, plug-and-play state update strategies TTT3R or TTSA3R, \modelname further yields better results compared to applying these strategies to CUT3R.}
\label{fig:cam_pose_err_long}
\end{figure*}

\subsection{Channel-wise Update of Explicit Memory}
\label{subsec:channel_wise}

To represent persistent global geometry, we retain an explicit token-based memory
    $S_t \in \mathbb{R}^{N_s \times C},$
where $N_s$ is the number of state tokens and $C$ is the channel dimension. The decoder output $S'_t$ in \cref{eq:decoder} is treated as a \emph{candidate} state update rather than the final new state.

Directly overwriting the state can lead to temporal forgetting in long sequences. To address this issue, we introduce a channel-wise update gate $\zeta_t \in (0,1)^{N_s \times C}$:
\begin{equation}
    \zeta_t = \sigma\!\left(f_{\zeta}(F_t, S_{t-1})\right),
    \label{eq:gate}
\end{equation}
where $f_{\zeta}$ is an MLP. The explicit memory is then updated as
\begin{equation}
    S_t
    =
    \zeta_t \odot S'_t
    +
    (1-\zeta_t) \odot S_{t-1}.
\end{equation}
This channel-wise gating enables finer control over memory evolution, allowing stable geometric features to be preserved while selectively integrating new observations.
During inference, when applying the per-token state-update mechanism of TTT3R or TTSA3R, we combine its per-token gate $G_t$ with our channel-wise gate $\zeta_t$ to obtain the final update gate. The state is updated as
\begin{equation}
    S_t = G_t \odot \zeta_t \odot S'_t
    + \left(1 - G_t \odot \zeta_t\right) \odot S_{t-1},
\end{equation}
where $G_t \in \mathbb{R}^{N_s \times 1}$ is broadcast along the channel dimension and controls token-level updates, while $\zeta_t \in (0,1)^{N_s \times C}$ controls channel-level updates. We provide more details on this module in the supplementary material.

\subsection{Prediction Heads and Training}
\label{subsec:heads}

We follow CUT3R in using the same prediction heads and supervision. Given the refined tokens $(p_t, F'_t)$, \modelname predicts local-frame pointmaps and confidence scores, $(\hat{Z}_t^{\mathrm{self}}, C_t^{\mathrm{self}})
    =
    \mathrm{Head}_{\mathrm{self}}(F'_t)$, world-frame pointmaps and confidence scores,
$
    (\hat{Z}_t^{\mathrm{world}}, C_t^{\mathrm{world}})
    =
    \mathrm{Head}_{\mathrm{world}}(p_t, F'_t),
$
and the 6-DoF camera pose
$
    \hat{P}_t = \mathrm{Head}_{\mathrm{pose}}(p_t).
$

For training, we use the same objectives as CUT3R. The local- and world-frame pointmaps are supervised with the confidence-aware 3D regression loss in CUT3R and MASt3R~\cite{leroy2024grounding}:

\begin{equation}
\mathcal{L}_{\mathrm{3D}}
=
\sum_{f\in\{\mathrm{self},\mathrm{world}\}}
\mathcal{L}_{\mathrm{conf}}
\!\left(\hat{Z}^{f}, Z^{f}, C^{f}\right).
\end{equation}

\begin{equation}
    \mathcal{L}_{\mathrm{conf}}(\hat{Z}, Z, C)
=
\sum_{(\hat{z},z, c)\in(\hat{Z}, Z, C)}
\left(
c \cdot \left\lVert \frac{\hat{z}}{\hat{s}} - \frac{z}{s} \right\rVert_2
-
\beta \log c
\right),
\label{eq:3d_loss}
\end{equation}
where $\hat{s}$ and $s$ are scale normalization factors for $\hat{Z}$ and ${Z}$, respectively.

Camera pose is supervised by
$
\mathcal{L}_{\mathrm{pose}}
=
\sum_{t=1}^{N}
\left(
\left\lVert \hat{q}_t - q_t \right\rVert_2
+
\left\lVert \frac{\hat{\tau}_t}{\hat{s}} - \frac{\tau_t}{s} \right\rVert_2
\right),
$
and, when raymap inputs are available, we additionally use the RGB reconstruction loss
$
\mathcal{L}_{\mathrm{rgb}} = \left\lVert \hat{I}_r - I_r \right\rVert_2^2.
$
The full objective is
$
\mathcal{L}
=
\mathcal{L}_{\mathrm{3D}}
+
\mathcal{L}_{\mathrm{pose}}
+
\mathbbm{1}_{\mathrm{raymap}} \mathcal{L}_{\mathrm{rgb}}, 
$
where $\mathbbm{1}_{\mathrm{raymap}}$ indicates whether the input is a raymap. 

To ensure that the gains can be fairly attributed to our architectural design, we train \modelname using the same 26-dataset mixture and training configuration as the final training stage of CUT3R, with input sequences ranging from 4 to 64 views. The training data cover diverse real and synthetic scenes, including CO3Dv2~\cite{reizenstein2021common}, ARKitScenes~\cite{baruch2021arkitscenes}, MegaDepth~\cite{li2018megadepth}, MapFree~\cite{arnold2022map}, DL3DV~\cite{ling2024dl3dv}, and Hypersim~\cite{roberts2021hypersim}. The full dataset list is provided in the supplementary material. We initialize the encoder, decoders, and prediction heads from pre-trained 512-DPT CUT3R weights and fine-tune all modules except the encoder. Optimization is performed with AdamW~\cite{loshchilov2017decoupled}, using an initial learning rate of $1\times10^{-6}$, linear warmup, and cosine decay. Training takes approximately 10 hours on 32 NVIDIA H100 GPUs with a batch size of 8 per GPU.

\section{Experiment}
\label{sec:exp}

\begin{figure*}[htbp]
\centering
\includegraphics[width=0.98\textwidth]{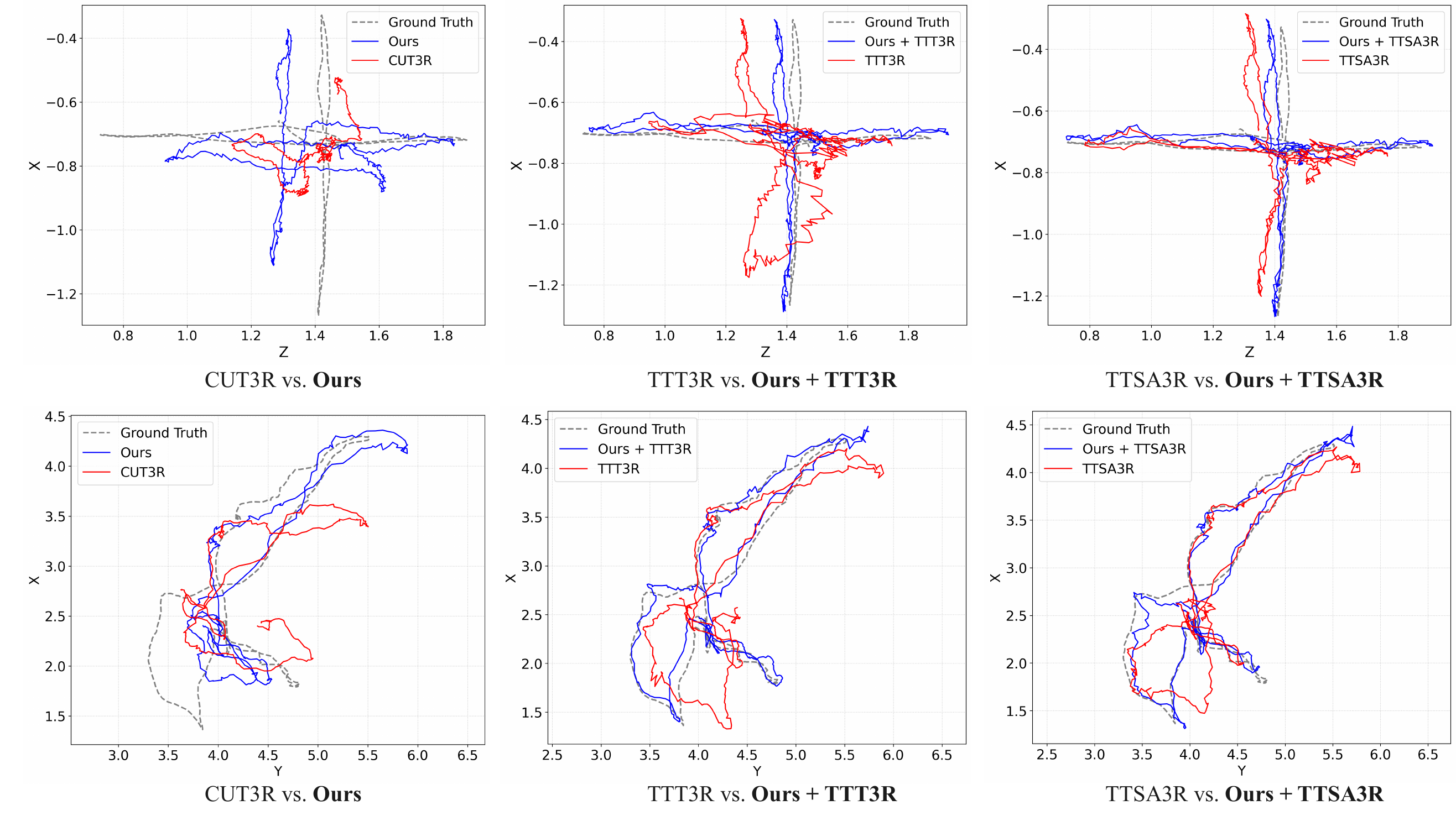}
\caption{Qualitative visualization of predicted camera trajectories on the TUM Dynamics dataset (top row) and the ScanNet dataset (bottom row). Each row is evaluated on the same sequence. Our \modelname achieves significant improvement compared with CUT3R. Moreover, when combined with TTT3R or TTSA3R, \modelname further reduces drift compared to applying these strategies to CUT3R.}

\label{fig:cam_pose_traj}
\end{figure*}

We evaluate our \modelname on a variety of tasks, including camera pose estimation (\cref{subsec:cam_pose_est}), video
depth estimation (\cref{subsec:video_depth_est}), and 3D reconstruction (\cref{subsec:3d_recon}). We  primarily focus on streaming 3D reconstruction models, specifically Point3R~\cite{wu2025point3r}, StreamVGGT~\cite{zhuo2025streaming}, and CUT3R~\cite{wang2025continuous}. Among these, we treat the RNN-based CUT3R as our closest baseline to demonstrate that \modelname achieves superior performance on long sequences ($\geq$ 200 frames). Furthermore, we show that \modelname is fully compatible with existing training-free state update mechanisms developed for CUT3R, such as TTT3R~\cite{chen2026tttr} and TTSA3R~\cite{zheng2026ttsa3r}. Quantitative accuracy results are obtained using NVIDIA A100 40GB GPUs. 
Qualitative visualizations and efficiency measurements are conducted separately on a workstation equipped with a single NVIDIA RTX PRO 6000 Blackwell Max-Q GPU.

\subsection{Camera Pose Estimation}
\label{subsec:cam_pose_est}
Following~\cite{chen2026tttr, zheng2026ttsa3r}, we evaluate camera pose accuracy on the TUM Dynamics~\cite{sturm2012benchmark} and ScanNet~\cite{scannet-DaiCSHFN17} datasets using sequences ranging from 50 to 1,000 frames. We adopt the standard Absolute Trajectory Error (ATE) metric, computed after Sim(3) alignment~\cite{umeyamaleast} between the estimated and ground-truth trajectories. As illustrated in~\cref{fig:cam_pose_err_long}, \modelname consistently outperforms CUT3R and Point3R. Furthermore, integrating TTT3R and TTSA3R into \modelname yields significantly lower pose errors, particularly on sequences exceeding 400 streaming frames. In contrast, VGGT and StreamVGGT, both based on full attention, are prone to memory exhaustion for such long sequences. Qualitative comparisons in~\cref{fig:cam_pose_traj} demonstrate that \modelname, when equipped with TTT3R and TTSA3R, achieves superior tracking accuracy compared to CUT3R under identical state-update strategies. 

Notably, from~\cref{fig:cam_pose_err_long}, \modelname with TTT3R provides a 39\% relative reduction in ATE at the 500-frame mark on the TUM dynamics dataset compared with the base TTT3R model. 
More results including short sequences are provided in the supplementary material.

\begin{table*}[htbp]
\centering
\caption{Quantitative evaluation of video depth estimation on long sequences from the KITTI dataset. \textbf{Bold} indicates the best result per column within each depth category, and \colorbox{lightgreen}{green} indicates improved or maintained performance of our model over its respective base counterpart.}
\resizebox{\textwidth}{!}{ 
\begin{tabular}{l cccccccccc}
\toprule
\multirow{2}{*}{\textbf{Method}} 
& \multicolumn{2}{c}{\textbf{300 frames}} 
& \multicolumn{2}{c}{\textbf{350 frames}} 
& \multicolumn{2}{c}{\textbf{400 frames}} 
& \multicolumn{2}{c}{\textbf{450 frames}} 
& \multicolumn{2}{c}{\textbf{500 frames}} \\
\cmidrule(lr){2-3} 
\cmidrule(lr){4-5} 
\cmidrule(lr){6-7} 
\cmidrule(lr){8-9} 
\cmidrule(lr){10-11}
& Abs Rel $\downarrow$ & $\delta_{1.25} \uparrow$
& Abs Rel $\downarrow$ & $\delta_{1.25} \uparrow$
& Abs Rel $\downarrow$ & $\delta_{1.25} \uparrow$
& Abs Rel $\downarrow$ & $\delta_{1.25} \uparrow$
& Abs Rel $\downarrow$ & $\delta_{1.25} \uparrow$ \\

\midrule
\multicolumn{11}{c}{\textbf{Metric Depth}} \\
\midrule

CUT3R
& 0.134 & 83.8
& 0.140 & 82.7
& 0.144 & 82.1
& 0.149 & 81.1
& 0.152 & 80.4 \\

\ours
& \gb 0.119 & \gb 88.4
& \gb 0.123 & \gb 87.4
& \gb 0.126 & \gb 87.1
& \gb 0.129 & \gb 86.8
& \gb 0.132 & \gb 86.0 \\

\midrule

TTT3R
& 0.117 & 88.5
& 0.122 & 87.7
& 0.125 & 87.3
& 0.130 & 86.6
& 0.132 & 86.5 \\

\ours + TTT3R
& \gb \textbf{0.112} & \gb \textbf{89.5}
& \gb \textbf{0.115} & \gb \textbf{88.9}
& \gb \textbf{0.119} & \gb \textbf{88.6}
& \gb \textbf{0.122} & \gb \textbf{88.2}
& \gb \textbf{0.125} & \gb \textbf{87.5} \\

\midrule

TTSA3R
& 0.117 & 88.8
& 0.121 & 88.0
& 0.126 & 87.5
& 0.131 & 86.7
& 0.133 & 86.5 \\

\ours + TTSA3R
& \gb 0.116 & \gb 89.0
& \gb 0.118 & \gb 88.5
& \gb 0.122 & \gb 88.2
& \gb 0.125 & \gb 87.7
& \gb 0.129 & \gb 87.0 \\

\midrule
\multicolumn{11}{c}{\textbf{Scale-Invariant Depth}} \\
\midrule

CUT3R
& 0.122 & 87.4
& 0.125 & 87.0
& 0.126 & 87.1
& 0.129 & 86.7
& 0.130 & 86.5 \\

\ours
& \gb 0.107 & \gb 90.3
& \gb 0.110 & \gb 89.6
& \gb 0.110 & \gb 89.4
& \gb 0.111 & \gb 89.2
& \gb 0.113 & \gb 89.1 \\

\midrule

TTT3R
& 0.112 & 90.2
& 0.114 & 89.6
& 0.115 & 89.5
& 0.118 & 89.1
& 0.119 & 89.1 \\

\ours + TTT3R
& \gb 0.104 & \gb 91.3
& \gb 0.106 & \gb 90.8
& \gb 0.107 & \gb 90.5
& \gb 0.109 & \gb 90.1
& \gb 0.110 & \gb 89.9 \\

\midrule

TTSA3R
& 0.109 & 90.9
& 0.111 & 90.2
& 0.113 & 89.9
& 0.116 & 89.5
& 0.117 & 89.4 \\

\ours + TTSA3R
& \gb \textbf{0.103} & \gb \textbf{92.0}
& \gb \textbf{0.104} & \gb \textbf{91.5}
& \gb \textbf{0.105} & \gb \textbf{91.2}
& \gb \textbf{0.108} & \gb \textbf{90.7}
& \gb \textbf{0.109} & \gb \textbf{90.5} \\

\bottomrule
\end{tabular}
}
\label{tab:depth_estimation_kitti_merged}
\end{table*}

\begin{table*}[htbp]
\centering
\caption{Quantitative evaluation on the Bonn dataset. \textbf{Bold} indicates the absolute best result per column within each depth category; \colorbox{lightgreen}{green} indicates improved or maintained performance over its direct baseline.}
\resizebox{\textwidth}{!}{ 
\begin{tabular}{l cccccccccc}
\toprule
\multirow{2}{*}{\textbf{Method}} 
& \multicolumn{2}{c}{\textbf{300 frames}} 
& \multicolumn{2}{c}{\textbf{350 frames}} 
& \multicolumn{2}{c}{\textbf{400 frames}} 
& \multicolumn{2}{c}{\textbf{450 frames}} 
& \multicolumn{2}{c}{\textbf{500 frames}} \\
\cmidrule(lr){2-3} 
\cmidrule(lr){4-5} 
\cmidrule(lr){6-7} 
\cmidrule(lr){8-9} 
\cmidrule(lr){10-11}
& Abs Rel $\downarrow$ & $\delta_{1.25} \uparrow$
& Abs Rel $\downarrow$ & $\delta_{1.25} \uparrow$
& Abs Rel $\downarrow$ & $\delta_{1.25} \uparrow$
& Abs Rel $\downarrow$ & $\delta_{1.25} \uparrow$
& Abs Rel $\downarrow$ & $\delta_{1.25} \uparrow$ \\

\midrule
\multicolumn{11}{c}{\textbf{Metric Depth}} \\
\midrule

CUT3R 
& 0.106 & 88.8
& 0.107 & 88.8
& 0.106 & 89.5
& 0.101 & 90.2
& 0.10 & 90.6 \\

\ours
& \gb 0.105 & \gb 89.7
& \gb 0.107 & \gb 89.6
& \gb 0.105 & \gb 90.2
& \gb 0.101 & \gb 90.8
& \gb 0.10 & \gb 91.2 \\

\midrule

TTT3R
& 0.108 & 90.2
& 0.105 & 90.8
& 0.103 & 91.3
& 0.102 & 91.8
& 0.10 & 92.1 \\

\ours + TTT3R
& \gb 0.105 & \gb 91.4
& \gb 0.101 & \gb 91.9
& \gb 0.100 & \gb 92.3
& \gb 0.099 & \gb 92.6
& \gb 0.098 & \gb 92.9 \\

\midrule

TTSA3R
& 0.103 & 91.7
& 0.101 & 92.1
& 0.100 & 92.5
& 0.099 & 92.7
& 0.098 & 93.0 \\

\ours + TTSA3R
& \gb \textbf{0.102} & \gb \textbf{92.5}
& \gb \textbf{0.099} & \gb \textbf{92.8}
& \gb \textbf{0.098} & \gb \textbf{93.2}
& \gb \textbf{0.096} & \gb \textbf{93.4}
& \gb \textbf{0.094} & \gb \textbf{93.7} \\

\midrule
\multicolumn{11}{c}{\textbf{Scale-Invariant Depth}} \\
\midrule

CUT3R
& 0.089 & 93.8
& 0.091 & 93.1
& 0.090 & 93.4
& 0.086 & 93.6
& 0.084 & 93.8 \\

\ours
& \gb 0.085 & \gb 94.6
& \gb 0.087 & \gb 94.0
& \gb 0.087 & \gb 94.1
& \gb 0.085 & \gb 94.4
& \gb 0.083 & \gb 94.5 \\

\midrule

TTT3R
& 0.079 & 94.9
& 0.078 & 95.0
& 0.078 & 95.1
& 0.077 & 95.2
& 0.076 & 95.3 \\

\ours + TTT3R
& \gb 0.077 & \gb 94.9
& \gb 0.076 & \gb 95.0
& \gb 0.076 & \gb 95.1
& \gb 0.076 & \gb 95.2
& \gb 0.075 & \gb 95.3 \\

\midrule

TTSA3R
& 0.078 & \textbf{95.1}
& 0.077 & \textbf{95.1}
& 0.077 & \textbf{95.2}
& 0.077 & \textbf{95.3}
& 0.076 & \textbf{95.4} \\

\ours + TTSA3R
& \gb \textbf{0.076} & 95.0
& \gb \textbf{0.074} & 95.0
& \gb \textbf{0.074} & 95.1
& \gb \textbf{0.074} & 95.2
& \gb \textbf{0.073} & 95.3 \\

\bottomrule
\end{tabular}
}
\label{tab:depth_estimation_bonn_merged}
\end{table*}

\subsection{Video Depth Estimation} 
\label{subsec:video_depth_est}
As in~\cite{zhang2025monstr, wang2025continuous}, we evaluate long-sequence video depth estimation on the KITTI~\cite{geiger2013vision} and Bonn~\cite{palazzolo2019refusion} datasets. These datasets span 300 to 500 frames and encompass a diverse range of dynamic, static, indoor, and outdoor scenes. We report both scale-invariant and metric-scale results using the Absolute Relative error (Abs Rel) and the $\delta_{1.25}$ threshold (the percentage of pixels where $\max(\frac{d_{gt}}{d_{est}}, \frac{d_{est}}{d_{gt}}) < 1.25$).  As demonstrated in \cref{tab:depth_estimation_kitti_merged,tab:depth_estimation_bonn_merged}, \modelname significantly outperforms CUT3R. Integrating \modelname with TTT3R and TTSA3R further yields performance gains across most scenarios. Furthermore, we evaluate short-sequence performance (50 streaming frames) on the Sintel~\cite{butler2012naturalistic} dataset in the supplementary material.

\begin{figure}[htbp]
\centering
\includegraphics[width=0.5\textwidth]{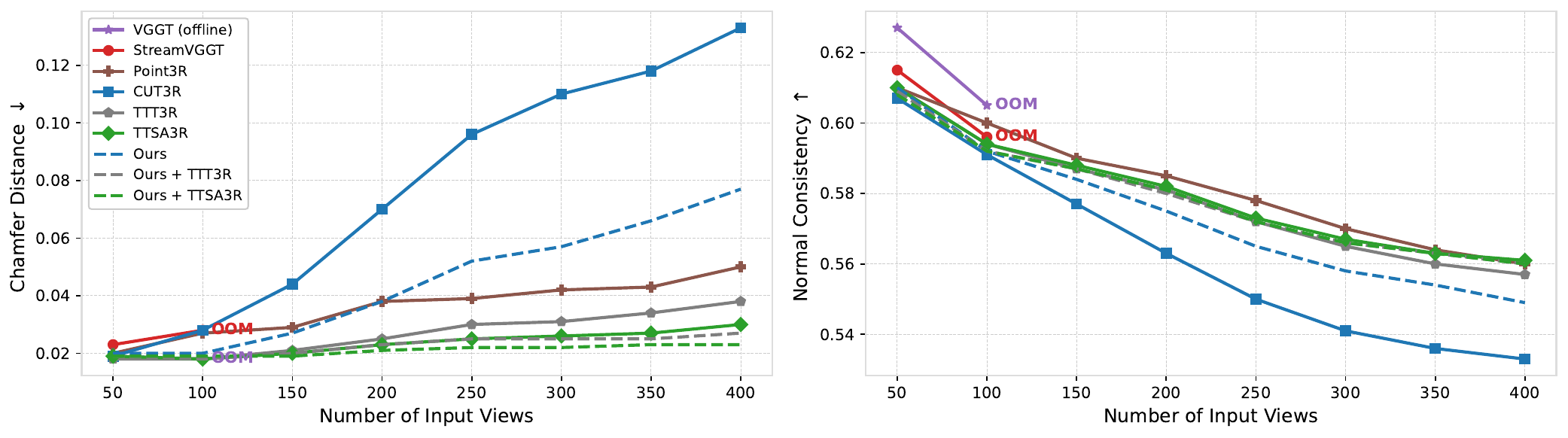}
\caption{Quantitative evaluation of 3D Reconstruction on 7-Scenes. OOM denotes out-of-memory. }
\label{fig:3d_recon_err_long}
\end{figure}

\begin{figure*}[htbp]
\centering
\includegraphics[width=0.88\textwidth]{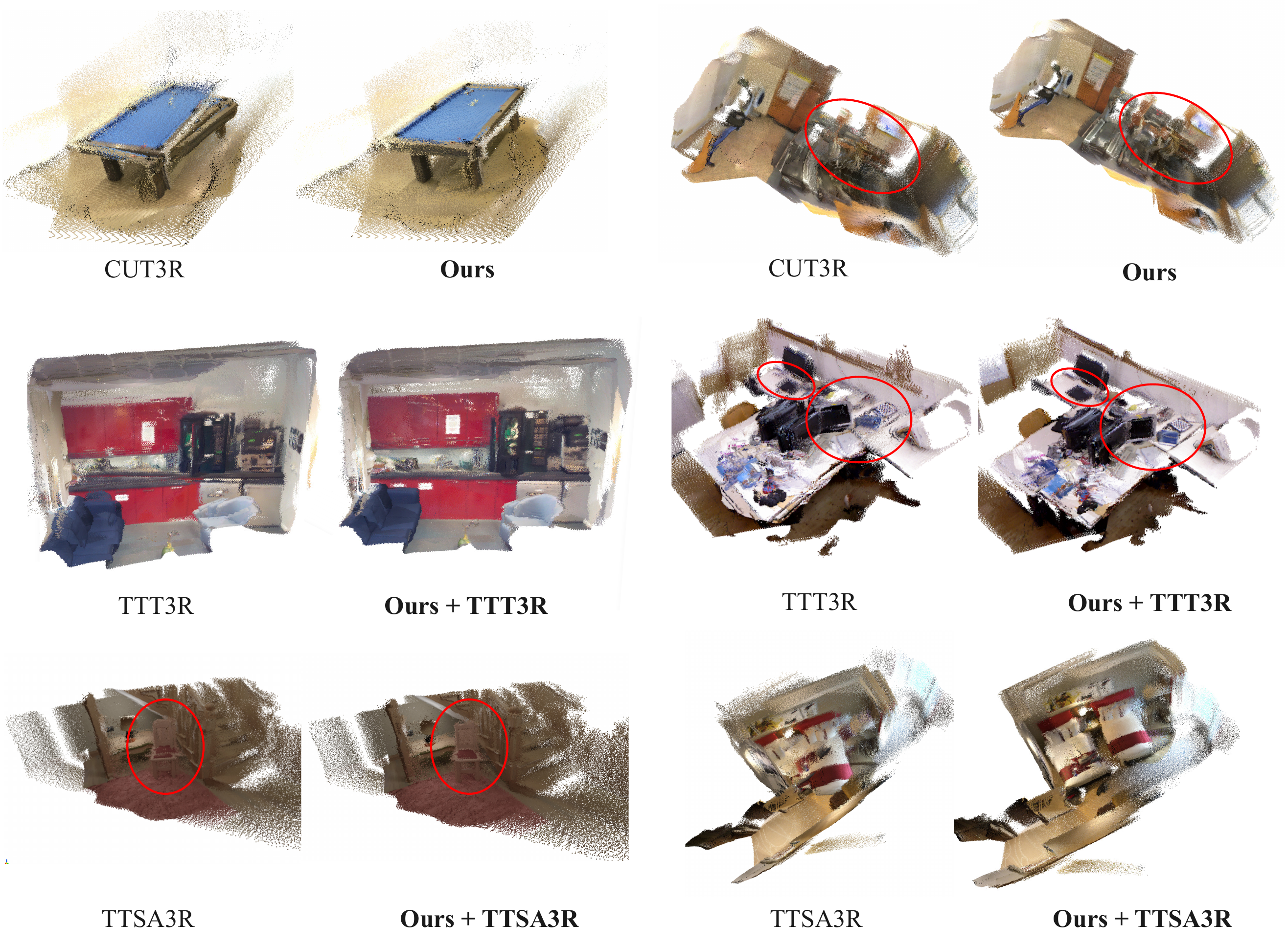}
\caption{Qualitative 3D reconstruction results on long sequences. \textcolor{red}{Red circles} mark some representative regions where the difference is particularly clear.}
\label{fig:vis_3drecon}
\end{figure*}

\subsection{3D Reconstruction}
\label{subsec:3d_recon}

We evaluate our multiview reconstruction framework on the 7-Scenes~\cite{shotton2013scene} dataset over sequences of 50 to 400 frames, using Chamfer Distance (CD) and Normal Consistency (NC) as the primary metrics. Following TTT3R~\cite{chen2026tttr}, CD is defined as the mean of accuracy and completeness, where accuracy measures the Euclidean distance from reconstructed points to the nearest ground-truth points and completeness is defined conversely. As shown in \cref{fig:3d_recon_err_long}, \modelname significantly outperforms CUT3R. Moreover, when combined with TTT3R or TTSA3R, \modelname consistently achieves lower CD than Point3R and CUT3R under the same state-update strategy, further validating the effectiveness of our architecture. Qualitative results in \cref{fig:vis_3drecon,fig:teaser} show that \modelname produces more accurate and better-aligned 3D reconstructions, with further improvements when combined with TTT3R or TTSA3R. For example, in \cref{fig:teaser}, \ours + TTT3R reconstructs the staircase with more accurate step spacing and clearer handrails. Additional visualizations and results on NRGBD~\cite{azinovic2022neural} are reported in the supplementary material.

\subsection{Model Efficiency}
\label{subsec:model_eff}
\modelname maintains constant GPU memory complexity with respect to sequence length. As shown in \cref{tab:runtime_memory}, \modelname matches the throughput of CUT3R and its training-free state update variants while reducing GPU memory usage by 7\% and parameter count by 19\% relative to CUT3R. All evaluations are conducted on a workstation with a single NVIDIA RTX PRO 6000 Blackwell Max-Q GPU using $512\times384$ images from 7-Scenes.

\begin{table}[ht]

\centering

\caption{Efficiency comparison of runtime (fps), GPU memory usage (MiB), and parameter count. \colorbox{lightgreen}{Green} indicates that our model matches or outperforms its corresponding base model.}

\resizebox{0.5\textwidth}{!}{

\begin{tabular}{lccc}

\toprule

\textbf{Method} & \textbf{Runtime (fps)} $\uparrow$ & \textbf{Memory (MiB)} $\downarrow$ & \textbf{Params} $\downarrow$\\

\midrule

CUT3R & 26 & 7930 & 793M\\

\ours & \gb 26 & \gb 7340 & \gb 644M \\

\midrule

TTT3R & 25 & 8364 & 793M\\

\ours + TTT3R & \gb 25 & \gb 7774 & \gb 644M\\

\midrule

TTSA3R & 25 & 8786 & 793M\\

\ours + TTSA3R & \gb 25 & \gb 8208 & \gb  644M\\

\bottomrule

\end{tabular}

}

\label{tab:runtime_memory}

\end{table}

\subsection{Ablation Study}
\label{subsec:abla}

\begin{table}[t]
\centering
\caption{
Ablation study of \modelname. Both variants are independently retrained after removing the corresponding module. 1. w/o. channel-wise gate replaces CUT3R's original 152M-parameter Transformer-based pose \texttt{LocalMemory} with our implicit fast-weight pose memory, while removing the proposed channel-wise state update mechanism. 2. w/o. fast weight uses CUT3R’s original 152M-parameter Transformer-based pose \texttt{LocalMemory} while retaining the proposed channel-wise update mechanism.
}
\resizebox{0.92\columnwidth}{!}{%
\begin{tabular}{lcccccc}
\toprule
\multicolumn{7}{c}{
\textbf{TUM Dynamics: ATE (m) $\downarrow$}
} \\
\midrule
\textbf{Method}
& \multicolumn{6}{c}{\textbf{\#frames}} \\
\cmidrule(lr){2-7}
& 200 & 400 & 600 & 800 & \multicolumn{2}{c}{1000} \\
\midrule
TTT3R
& 0.041 & 0.051 & 0.074 & 0.091
& \multicolumn{2}{c}{0.110} \\
\ours + TTT3R (w/o. gate)
& 0.031 & 0.044 & 0.059 & 0.090
& \multicolumn{2}{c}{0.100} \\
\ours + TTT3R (w/o. fast weight)
& 0.027 & 0.039 & 0.054 & 0.067
& \multicolumn{2}{c}{0.078} \\
\ours + TTT3R
& \textbf{0.025} & \textbf{0.037}
& \textbf{0.046} & \textbf{0.060}
& \multicolumn{2}{c}{\textbf{0.073}} \\
\midrule
\multicolumn{7}{c}{\textbf{KITTI: video depth}} \\
\midrule
\textbf{Method}
& \multicolumn{2}{c}{\textbf{300 frames}}
& \multicolumn{2}{c}{\textbf{400 frames}}
& \multicolumn{2}{c}{\textbf{500 frames}} \\
\cmidrule(lr){2-3}
\cmidrule(lr){4-5}
\cmidrule(lr){6-7}
& Abs Rel $\downarrow$ & $\delta_{1.25} \uparrow$
& Abs Rel $\downarrow$ & $\delta_{1.25} \uparrow$
& Abs Rel $\downarrow$ & $\delta_{1.25} \uparrow$ \\
\midrule
TTT3R
& 0.11 & 90.2 & 0.12 & 89.5 & 0.12 & 89.1 \\
\ours + TTT3R (w/o. gate)
& 0.11 & 91.0 & \textbf{0.11} & 90.0
& \textbf{0.11} & 89.5 \\
\ours + TTT3R (w/o. fast weight)
& 0.11 & 91.0 & \textbf{0.11} & 90.0 & 0.12 & 89.5 \\
\ours + TTT3R
& \textbf{0.10} & \textbf{91.3}
& \textbf{0.11} & \textbf{90.5}
& \textbf{0.11} & \textbf{89.9} \\
\bottomrule
\end{tabular}%
}
\label{tab:abl_ttt3r_combined}
\end{table}

To evaluate the contributions of the components introduced in \cref{subsec:fast_weight,subsec:channel_wise}, we perform ablation studies on long-sequence camera pose estimation on the TUM Dynamics dataset and video depth estimation on the KITTI dataset, using TTT3R as the baseline. As shown in \cref{tab:abl_ttt3r_combined}, both the fast-weight memory and the channel-wise state gate help mitigate temporal forgetting in streaming 3D perception, leading to better accuracy than the original TTT3R.

\section{Conclusion}
\textbf{\modelname} is an RNN-based hybrid memory model for streaming 3D reconstruction that decouples camera tracking from geometric mapping, improving long-term consistency and long-sequence performance while preserving recurrent streaming efficiency. Compared with CUT3R, \modelname is also more parameter-efficient, reducing the model size from 793M to 644M parameters. Despite being more compact, it significantly lowers the Absolute Trajectory Error (ATE) on the challenging 500-1000 frame benchmark of the TUM Dynamics and ScanNet datasets. The resulting improvements also extend to downstream tasks such as 3D reconstruction and video depth estimation. These results establish hybrid memory as an effective design for accurate and efficient long-horizon 3D perception.

\section*{Acknowledgments}
We would like to thank Guangyao Zhai for the valuable discussion during the initial stages of this project, and Haian Jin for the assistance with the training setup. We thank the members of Android XR, Google for their help and support, especially Xiao Yuan, Luke Jia and Chao Guo.

\section{Supplementary Material}

\subsection{Design Insight and Architectural Novelty}
\label{app:design_rationale}

Mem3R is not a direct combination of CUT3R~\cite{wang2025continuous} and test-time training. Existing training-free adaptation approaches typically retain the original CUT3R architecture and modify only the inference-time state-update rule. In contrast, Mem3R redesigns and retrains the recurrent memory architecture itself. The key design question is not simply whether test-time adaptation can be introduced, but rather \emph{which part of the recurrent state should be represented by an adaptive implicit memory, which information should remain explicitly stored, and how these different memory components should be updated}.

Our design is motivated by the different characteristics of camera pose and dense scene geometry. Camera pose is a compact $6$-DoF state with strong temporal continuity, whereas dense geometry is high-dimensional and must preserve surface structure, depth, local geometric details, and cross-view correspondences accumulated over time. CUT3R maintains an explicit geometric state and a separate Transformer-based \texttt{LocalMemory} for pose-related information. We replace the latter with a lightweight fast-weight memory while retaining the explicit geometric state.

Based on this observation, Mem3R decomposes the recurrent memory according to the information being represented. We selectively replace only the pose-related \texttt{LocalMemory} in CUT3R with a lightweight fast-weight memory, in which pose-related temporal information is represented implicitly through rapidly adapted model parameters. In contrast, we retain an explicit token-based state for dense geometry, since geometric information must preserve rich spatial structure accumulated across frames. We further introduce a channel-wise state update gate to regulate how newly observed geometric features are incorporated into the persistent geometric state.

This leads to three central architectural design choices:
\begin{enumerate}
    \item \textbf{Which recurrent state should use test-time adaptive memory.}
    We apply fast-weight adaptation specifically to the pose-related state rather than replacing the entire recurrent memory.

    \item \textbf{How different information should be represented.}
    Pose is represented by a compact implicit adaptive memory, while dense geometry remains explicitly represented by state tokens.

    \item \textbf{How persistent geometric memory should be updated.}
    Instead of uniformly overwriting or updating the geometric state, the proposed channel-wise gate adaptively controls information retention and incorporation at the feature-channel level.
\end{enumerate}

The ablation results in the main paper support these design choices. Removing either the fast-weight pose memory or the channel-wise state update gate consistently degrades performance, while combining both components yields the strongest long-horizon results. Moreover, as analyzed in \cref{app:arch}, Mem3R reduces the total parameter count from approximately $793$M to $644$M. Therefore, the observed improvement does not arise from increasing model capacity; instead, it results from using a more suitable memory representation and update mechanism for the different types of recurrent state.

Overall, the contribution of Mem3R is not the generic application of test-time training to an existing recurrent reconstruction model, but a structured redesign of its memory architecture. By separating low-dimensional pose dynamics from high-dimensional geometric state, and assigning each an appropriate representation and update mechanism, Mem3R achieves both improved long-horizon performance and substantially reduced model complexity.

\subsection{Model Details}
\label{app:arch}

The explicit geometric memory consists of $768$ state tokens, each with a feature dimension of $768$. We set the confidence regularization coefficient $\beta$ in \cref{eq:3d_loss} to $0.2$. In this section, we provide additional architectural details of the lightweight fast-weight implicit memory introduced in \cref{subsec:fast_weight} and the channel-wise state update module described in \cref{subsec:channel_wise}. We further analyze the source of the parameter reduction compared with the original pose-related \texttt{LocalMemory} module in CUT3R~\cite{wang2025continuous}.

\paragraph{Fast-weight memory via TTT.}
The fast-weight module serves as the core implicit memory for pose tracking and contains approximately $1.56$M parameters. It first projects the $1024$-dimensional visual features into a $768$-dimensional latent space, which is evenly divided into $12$ heads with a head dimension of $64$. The memory is implemented using a SwiGLU-based parameterization
$\mathcal{W}=\{W_1,W_2,W_3\}$, where
$W_1,W_2,W_3 \in \mathbb{R}^{64 \times 64}$ for each head.

Its online update is controlled by two lightweight linear predictors. The learning-rate head outputs $36$ scalars, corresponding to three learning rates for each of the $12$ heads, while the decay head predicts $12$ head-wise retention factors $\alpha$. During readout, the retrieved features are processed by an RMSNorm layer followed by a linear projection back to the output feature space. This design represents the pose-related temporal state directly in the model parameters rather than maintaining a large set of pose-memory tokens.

\paragraph{Channel-wise state update module.}
The channel-wise state update module regulates how new observations are incorporated into the explicit geometric state and contains approximately $0.98$M parameters. Specifically, it concatenates the $768$-dimensional historical state feature with the $1024$-dimensional current visual feature, resulting in a $1792$-dimensional input. The concatenated feature is projected to a bottleneck dimension of $384$, followed by a GELU activation and a second linear projection to $768$ dimensions. A sigmoid function is then applied element-wise to produce the channel-wise update gate, allowing different feature channels to adaptively retain historical information or absorb new observations.

\paragraph{Source of the parameter reduction.}
The parameter reduction primarily comes from replacing the entire pose-related \texttt{LocalMemory} module of CUT3R. In the CUT3R configuration used in our experiments, the encoder and decoder feature dimensions are
$\texttt{enc\_embed\_dim}=1024$ and
$\texttt{dec\_embed\_dim}=768$, respectively, with
$\texttt{local\_mem\_size}=256$.
The \texttt{LocalMemory} first projects the $1024$-dimensional encoder feature to $768$ dimensions and concatenates it with a $768$-dimensional pose feature. Consequently, each \texttt{LocalMemory} decoder block operates at a feature dimension of
\begin{equation}
    d_L = 2 \times 768 = 1536.
\end{equation}
CUT3R employs two Transformer blocks for memory writing and another two blocks for memory readout, resulting in four \texttt{LocalMemory} decoder blocks in total.

For each decoder block, the self-attention projections contain approximately
$4d_L^2 \approx 9.44$M parameters, and the cross-attention projections contribute another
$4d_L^2 \approx 9.44$M parameters. The feed-forward network uses a hidden dimension of $4d_L$ and therefore contains approximately
$8d_L^2 \approx 18.87$M parameters. The dominant parameter count of each block is thus
\begin{equation}
    4d_L^2 + 4d_L^2 + 8d_L^2
    = 16d_L^2
    \approx 37.75\text{M},
\end{equation}
or approximately $37.78$M when biases and normalization parameters are included. The four memory-writing and memory-readout blocks therefore account for approximately $151.12$M parameters. Together with the additional query projection, the learnable local memory of size $256\times1536$, and the masked token, the complete pose-related \texttt{LocalMemory} module contains approximately $152.31$M parameters.

\paragraph{Overall complexity.}
Our design removes this $152.31$M-parameter pose-related \texttt{LocalMemory} module and replaces it with the fast-weight pose memory and the channel-wise state update module, which contain approximately $1.56$M and $0.98$M parameters, respectively. The two proposed modules therefore introduce only approximately
\begin{equation}
    1.56\text{M} + 0.98\text{M} = 2.54\text{M}
\end{equation}
parameters in total. This yields a net parameter reduction of approximately
\begin{equation}
    152.31\text{M} - 2.54\text{M}
    = 149.77\text{M},
\end{equation}
which is consistent with the reduction of the full model from approximately $793$M to $644$M parameters. These results also support our design motivation: using multiple Transformer decoder blocks together with token-based storage is unnecessarily parameter-intensive for representing a compact $6$-DoF camera pose state. By replacing this pose-specific token memory with fast weights, our model substantially reduces parameter and memory overhead while achieving stronger long-horizon streaming 3D reconstruction and camera pose estimation performance.

\begin{figure*}[h!]
\centering
\includegraphics[width=\textwidth]{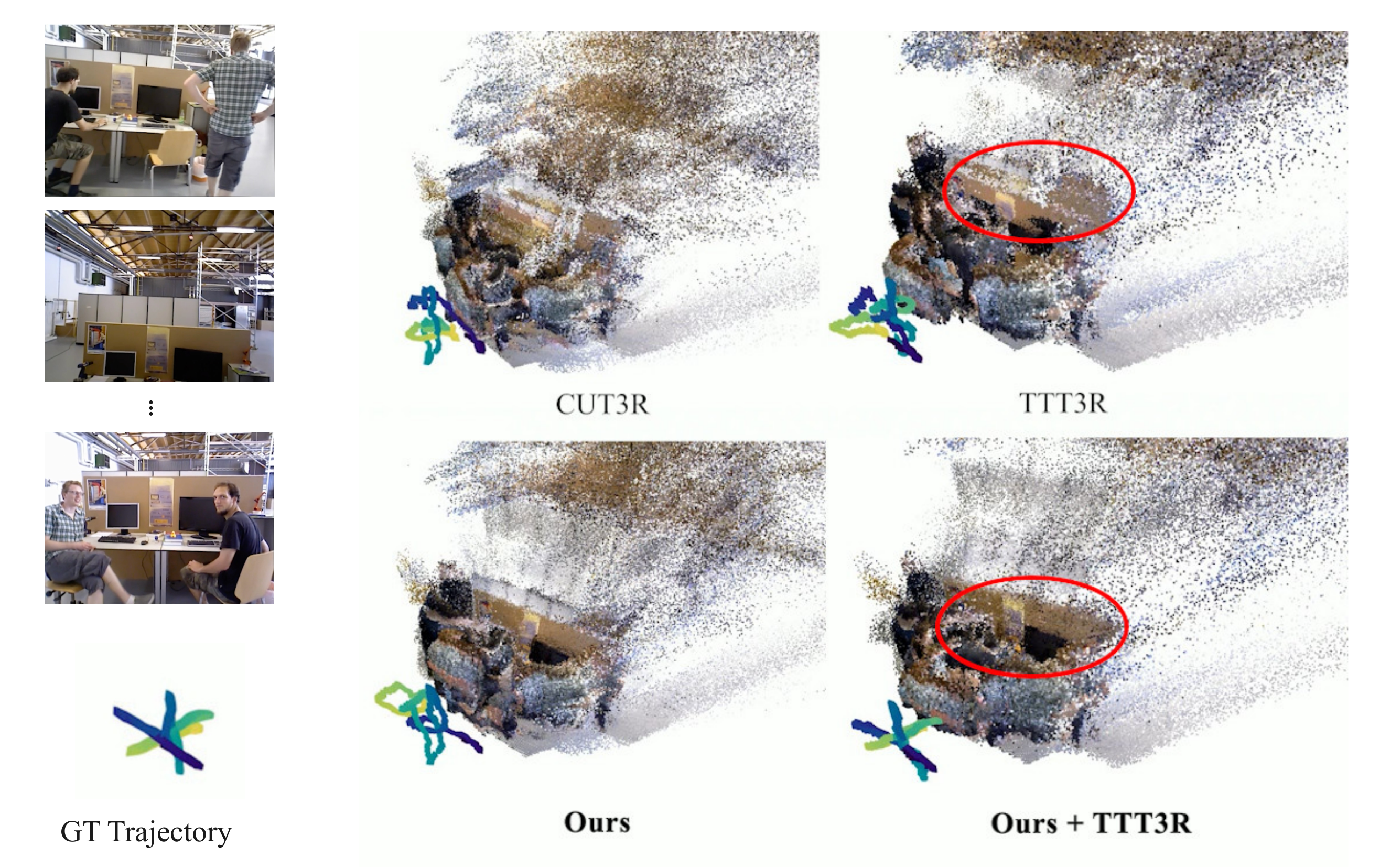}
\caption{Qualitative 3D reconstruction results on long sequences. \textcolor{red}{Red circles} mark some representative regions where the difference is particularly clear.}
\label{fig:app_vis_3drecon_tumd}
\end{figure*}

\begin{figure*}[h!]
\centering
\includegraphics[width=\textwidth]{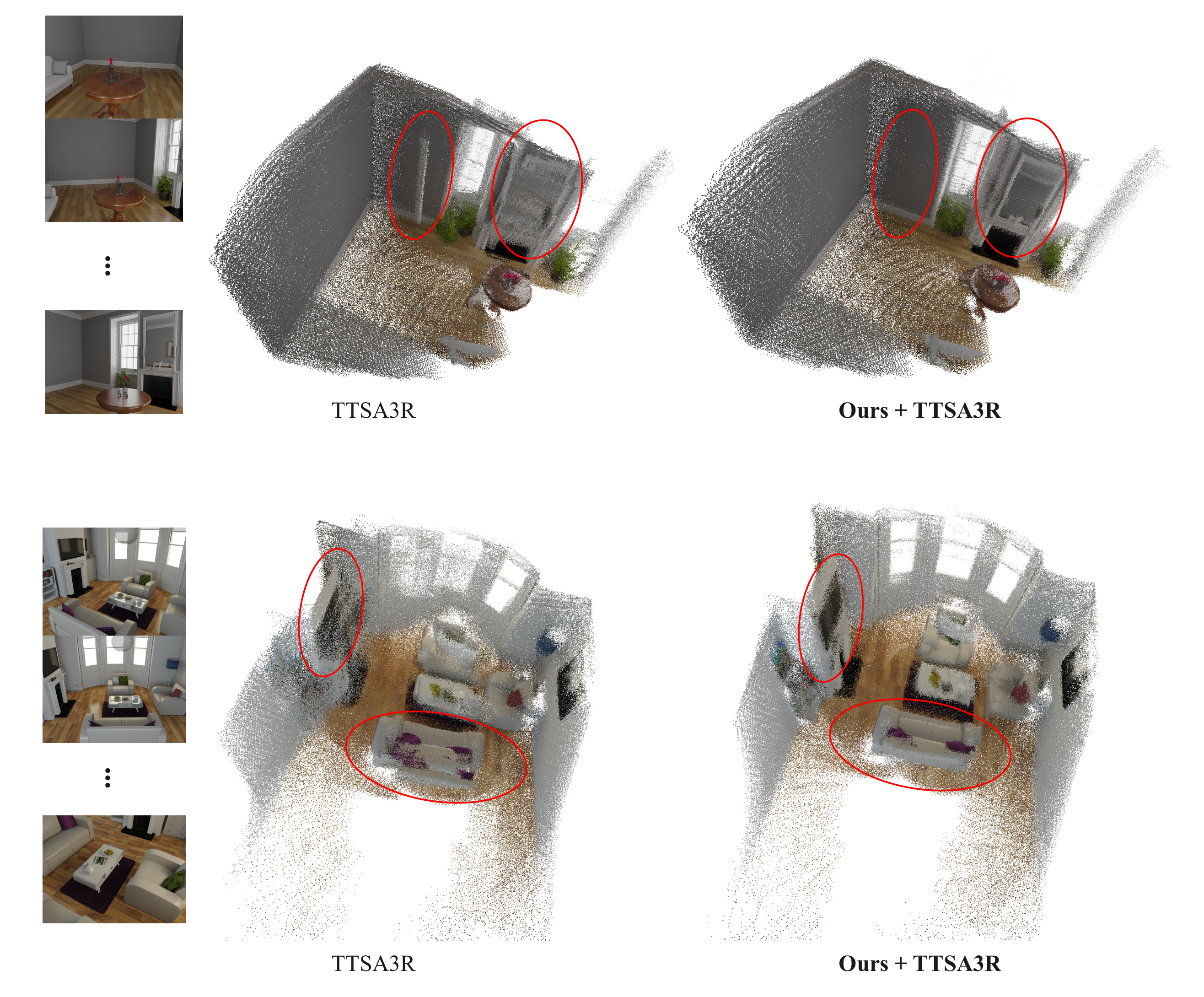}
\caption{Qualitative 3D reconstruction results on long sequences. \textcolor{red}{Red circles} mark some representative regions where the difference is particularly clear.}
\label{fig:app_vis_3drecon}
\end{figure*}

\begin{figure*}[htbp]
\centering
\includegraphics[width=\textwidth]{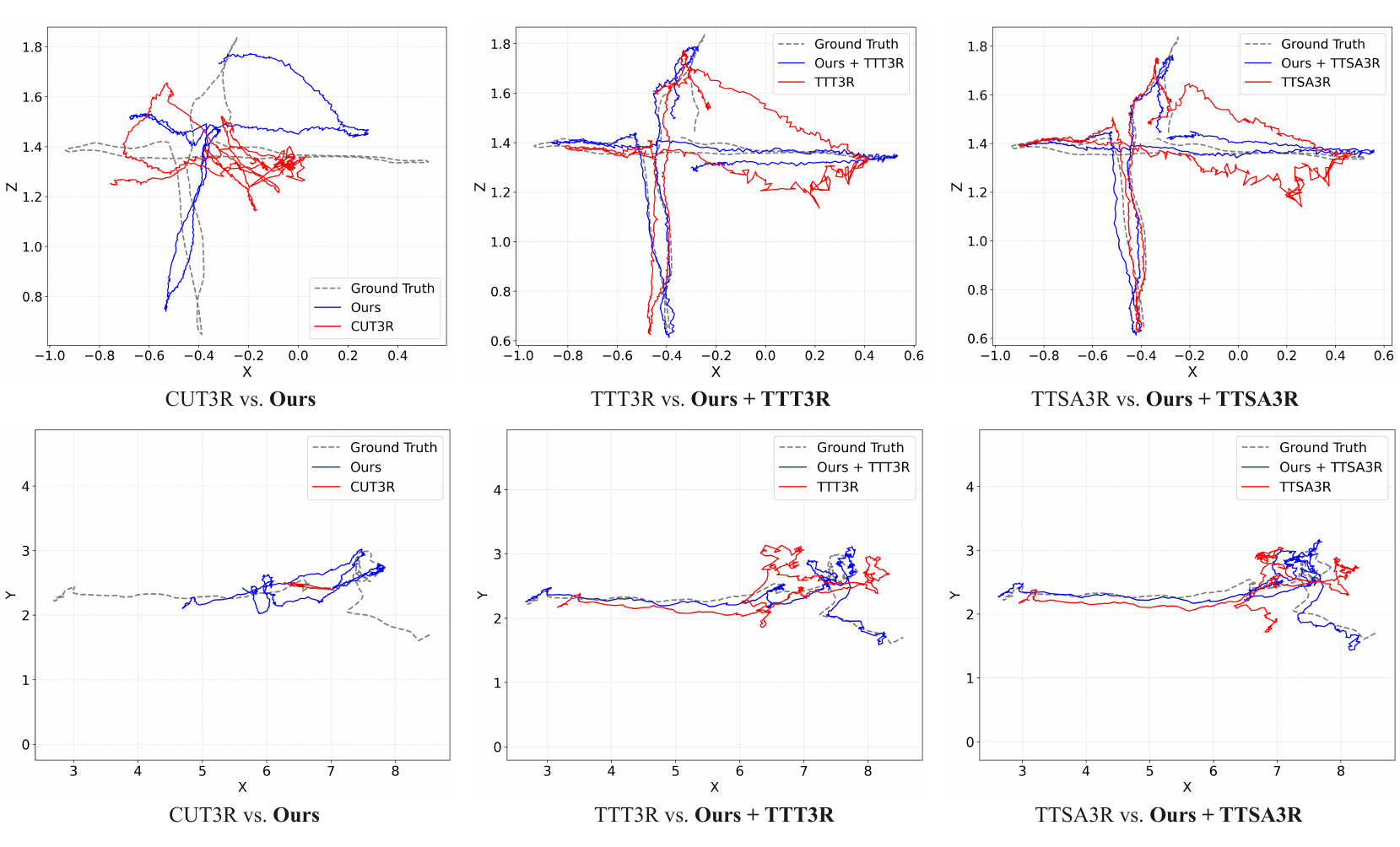}
\caption{Qualitative visualization of predicted camera trajectories on the TUM Dynamics dataset (top row) and the ScanNet dataset (bottom row). Each row is evaluated on the same sequence. Our \modelname achieves significant improvement compared with CUT3R. Moreover, when combined with TTT3R or TTSA3R, \modelname further reduces drift compared to applying these strategies to CUT3R.}

\label{fig:cam_pose_traj_sup}
\end{figure*}

\subsection{Supplement Visualization}
\label{app:sup_vis}
We provide additional qualitative results in \cref{fig:app_vis_3drecon_tumd,fig:app_vis_3drecon} to further compare the 3D reconstruction quality of \modelname against CUT3R, TTT3R, and TTSA3R. As discussed in \cref{subsec:3d_recon}, TTSA3R already improves the reconstruction quality of CUT3R substantially. Nevertheless, \cref{fig:app_vis_3drecon} shows that, under long streaming input sequences, \modelname combined with TTSA3R still produces more accurate reconstructions.

\subsection{Supplement Experiment}
\paragraph{Pose Estimation.}
\label{app:pose_est}
Additional qualitative comparisons in \cref{fig:cam_pose_traj_sup} show that, under the same state-update strategy, \modelname equipped with TTT3R or TTSA3R achieves more accurate camera tracking than CUT3R on long sequences.

\cref{tab:short_camera_pose_depth_sintel} shows that \modelname attains the lowest ATE among existing online 3D reconstruction models on Sintel~\cite{butler2012naturalistic} short sequences. At the same time, results in \cref{tab:ate_long_sequences_combined,tab:short_camera_pose_depth_sintel} suggest that incorporating TTT3R or TTSA3R into \modelname does not lead to substantial additional gains on short sequences. Given that \modelname has approximately 19\% fewer parameters than CUT3R while delivering clear improvements on long sequences, these short-sequence results still demonstrate its strong competitiveness.

\begin{table*}[ht]
    \centering
    \caption{ATE (m) ($\downarrow$) on 1,000-image sequences for ScanNet and TUM Dynamics. We evaluate \modelname{} equipped with TTT3R across various memory update strides.}
    \begin{tabular}{lcccccc}
        \toprule
        \multirow{2}{*}{Dataset} & \multicolumn{6}{c}{Memory Update Interval} \\
        \cmidrule(lr){2-7}
        & 1 & 2 & 4 & 6 & 8 & 10 \\
        \midrule
        ScanNet & 0.29 & \textbf{0.25} & 0.29 & 0.37 & 0.43 & 0.47 \\
        TUM Dynamics   & 0.073 & \textbf{0.056} & 0.071 & 0.077 & 0.095 & 0.10 \\
        \bottomrule
    \end{tabular}
    \label{tab:memory-intervals}

\end{table*}

\paragraph{Hybrid Memory Update Interval}
Given that \modelname is trained on sequences of up to 64 views, we investigate how the model adapts to significantly longer sequences during inference by varying the memory update interval $n \in \{1, 2, 4, 6, 8, 10\}$. We update the implicit and explicit memories only every $n$-th frame, whereas the ATE is computed across all frames.  Frames between two memory updates still produce pose predictions, but they are not used to update either memory. While our primary results in the main paper use a default stride of $n=1$, we find that an interval of $n=2$ achieves peak performance on the long sequences of TUM Dynamics and ScanNet, as shown in \cref{tab:memory-intervals}. Larger intervals eventually lead to a decline in accuracy due to the loss of fine-grained temporal information. This trend suggests that less frequent memory updates may improve long-sequence stability by reducing sensitivity to high-frequency frame-level noise and redundant observations. Updating the memory at every frame exposes the model to such noise and redundancy, whereas a moderate stride such as $n=2$ smooths the update process while retaining sufficient temporal resolution for accurate tracking.

\paragraph{Effect of Training Sequence Length} \label{app:training_length} We further investigate the effect of training sequence length on long-horizon generalization. In our main experiments, we intentionally follow the same training configuration as CUT3R, using sequences of $4$--$64$ views, together with the same data mixture, initialization, prediction heads, supervision, and optimization settings. This controlled setup ensures that the performance gains observed on long sequences are attributable to the proposed architecture rather than to a longer training context. To study the influence of the training horizon, we additionally retrain our model with maximum sequence lengths of $16$, $32$, $64$, and $96$ views, and evaluate relative camera pose estimation on ScanNet using sequences ranging from $50$ to $1000$ frames. The results are reported in \cref{tab:training_sequence_length}. Lower ATE is better, and the best result in each column is highlighted in \textbf{bold}.

\begin{table}[t] \centering \caption{ Effect of the maximum training sequence length on long-horizon relative camera pose estimation on ScanNet. We report ATE $\downarrow$ at different evaluation sequence lengths. Best results are shown in \textbf{bold}. } \resizebox{\columnwidth}{!}{ \begin{tabular}{lcccccccc} \toprule \textbf{Training length} & \textbf{50} & \textbf{100} & \textbf{150} & \textbf{200} & \textbf{400} & \textbf{600} & \textbf{800} & \textbf{1000} \\ \midrule Ours, 16 views & 0.042 & 0.085 & 0.141 & 0.216 & 0.518 & 0.685 & 0.740 & 0.782 \\ Ours, 32 views & \textbf{0.039} & 0.083 & \textbf{0.136} & \textbf{0.196} & 0.466 & 0.600 & 0.654 & 0.690 \\ Ours, 64 views & 0.040 & 0.083 & 0.138 & 0.200 & 0.432 & 0.584 & 0.650 & 0.706 \\ Ours, 96 views & \textbf{0.039} & \textbf{0.082} & 0.142 & 0.207 & \textbf{0.421} & \textbf{0.569} & \textbf{0.635} & \textbf{0.688} \\ \bottomrule \end{tabular} } \label{tab:training_sequence_length} \end{table} 

Longer training sequences generally improve long-horizon performance, particularly when evaluated on sequences of $400$--$1000$ frames. However, the trend is not strictly monotonic across all evaluation lengths. For example, the model trained with $32$ views performs best at $150$ and $200$ frames, whereas the $96$-view model achieves the best performance on most longer sequences. At $1000$ frames, increasing the maximum training length from $16$ to $96$ views reduces ATE from $0.782$ to $0.688$. These results indicate that increasing the training horizon improves generalization to long sequences, but does not by itself explain the gains of our architecture. Importantly, our main comparison with CUT3R uses the same maximum training horizon of $64$ views. Therefore, the improvements reported in the main paper are obtained under a controlled training setup and isolate the contribution of the proposed memory architecture.

\paragraph{Video Depth Estimation.}
\label{app:video_depth}
As shown in \cref{tab:short_camera_pose_depth_sintel,tab:video_depth_kitti_bonn_short}, \modelname achieves performance comparable to other online methods on short sequences. In line with TTT3R~\cite{chen2026tttr}, equipping \modelname with TTT3R or TTSA3R brings only limited improvements in this setting. Nevertheless, \modelname remains competitive with CUT3R on short sequences while reducing the parameter count by approximately 19\% and delivering substantial gains on long sequences. These results highlight the overall competitiveness of \modelname, particularly for the long-horizon streaming scenarios targeted by our design. 

\begin{table*}[!ht]
\centering
\caption{ATE (m) ($\downarrow$) of relative camera pose estimation on \textit{long sequences} on TUM Dynamics and ScanNet. \textbf{Bold} denotes the best result in each column within each dataset, and \colorbox{lightgreen}{green} indicates that our model matches or outperforms its corresponding base model.}
\resizebox{\linewidth}{!}{
\begin{tabular}{lcccccccccccc}
\toprule
\#frames & 50 & 100 & 150 & 200 & 300 & 400 & 500 & 600 & 700 & 800 & 900 & 1000 \\
\midrule
\multicolumn{13}{c}{\textbf{TUM Dynamics}} \\
\midrule
CUT3R  & 0.023 & 0.033 & 0.043 & 0.061 & 0.093 & 0.12 & 0.14 & 0.14 & 0.15 & 0.17 & 0.17 & 0.17 \\
\ours & \gb 0.012 & \gb 0.020 & \gb 0.031 & \gb 0.042 & \gb 0.064 & \gb 0.078 & \gb 0.096 & \gb 0.11 & \gb 0.13 & \gb 0.14 & \gb 0.15 & \gb 0.15 \\
\midrule
TTT3R & 0.014 & 0.020 & 0.031 & 0.041 & 0.045 & 0.051 & 0.067 & 0.074 & 0.085 & 0.091 & 0.102 & 0.11 \\
\ours + TTT3R & \gb \textbf{0.011} & \gb \textbf{0.015} & \gb 0.019 & \gb 0.025 & \gb 0.031 & \gb 0.037 & \gb 0.041 & \gb 0.046 & \gb 0.053 & \gb 0.060 & \gb 0.066 & \gb 0.073 \\
\midrule
TTSA3R & 0.012 & 0.018 & 0.025 & 0.030 & 0.038 & 0.043 & 0.056 & 0.064 & 0.074 & 0.079 & 0.086 & 0.091 \\
\ours + TTSA3R & \gb \textbf{0.011} & \gb \textbf{0.015} & \gb \textbf{0.018} & \gb \textbf{0.023} & \gb \textbf{0.027} & \gb \textbf{0.032} & \gb \textbf{0.036} & \gb \textbf{0.039} & \gb \textbf{0.045} & \gb \textbf{0.051} & \gb \textbf{0.058} & \gb \textbf{0.064} \\
\midrule
\multicolumn{13}{c}{\textbf{ScanNet}} \\
\midrule
CUT3R & 0.045 & 0.11 & 0.21 & 0.32 & 0.47 & 0.57 & 0.66 & 0.71 & 0.73 & 0.76 & 0.79 & 0.82 \\
\ours & \gb 0.040 & \gb 0.083 & \gb 0.14 & \gb 0.20 & \gb 0.33 & \gb 0.43 & \gb 0.53 & \gb 0.58 & \gb 0.62 & \gb 0.65 & \gb 0.67 & \gb 0.71 \\
\midrule
TTT3R & \textbf{0.033} & 0.072 & 0.11 & 0.14 & 0.19 & 0.24 & 0.28 & 0.32 & 0.34 & 0.37 & 0.40 & 0.40 \\
\ours + TTT3R & 0.034 & \gb 0.068 & \gb 0.10 & \gb 0.13 & \gb 0.17 & \gb 0.20 & \gb 0.22 & \gb \textbf{0.23} & \gb 0.25 & \gb 0.27 & \gb \textbf{0.28} & \gb \textbf{0.29} \\
\midrule
TTSA3R & \textbf{0.033} & \textbf{0.063} & 0.10 & 0.13 & 0.17 & 0.21 & 0.26 & 0.30 & 0.33 & 0.36 & 0.38 & 0.39 \\
\ours + TTSA3R & 0.035 & 0.065 & \gb \textbf{0.09} & \gb \textbf{0.12} & \gb \textbf{0.16} & \gb \textbf{0.18} & \gb \textbf{0.21} & \gb \textbf{0.23} & \gb \textbf{0.24} & \gb \textbf{0.26} & \gb \textbf{0.28} & \gb 0.30 \\
\bottomrule
\end{tabular}
}
\label{tab:ate_long_sequences_combined}
\end{table*}

\begin{table*}[ht]
\centering
\caption{Quantitative evaluation of camera pose and depth estimation on \textbf{short} sequences from the Sintel~\cite{butler2012naturalistic} dataset. For depth estimation, we report results under both per-sequence scale (\textit{Per-seq.}) and metric-scale settings, when available. GA denotes global alignment. `--` indicates that metric-scale depth is unavailable for methods that do not support metric-scale estimation. Best results are shown in \textbf{bold}.}
\resizebox{0.9\linewidth}{!}{
\begin{tabular}{l c ccc |cc cc}
\toprule
& & \multicolumn{7}{c}{\textbf{Sintel (50 frames)}} \\
\cmidrule(lr){3-9}
\textbf{Method} & \textbf{Online} 
& \multicolumn{3}{c}{\textbf{Camera pose}} 
& \multicolumn{2}{c}{\textbf{Depth (Per-seq.)}} 
& \multicolumn{2}{c}{\textbf{Depth (Metric)}} \\
\cmidrule(lr){3-5} \cmidrule(lr){6-7} \cmidrule(lr){8-9}
& & ATE $\downarrow$ & RPE trans $\downarrow$ & RPE rot $\downarrow$
& Abs Rel $\downarrow$ & $\delta_{1.25} \uparrow$
& Abs Rel $\downarrow$ & $\delta_{1.25} \uparrow$ \\
\midrule
DUSt3R-GA~\cite{DUSt3R}                 & \textcolor{red}{\ding{55}}  & 0.417 & 0.250 & 5.796 & 0.656 & 45.2 & -- & -- \\
MASt3R-GA~\cite{leroy2024grounding}     & \textcolor{red}{\ding{55}}  & 0.185 & 0.060 & 1.496 & 0.641 & 43.9 & \textbf{1.02} & \textbf{14.3}  \\
MonST3R-GA~\cite{zhang2025monstr}       & \textcolor{red}{\ding{55}}  & 0.111 & 0.044 & 0.869 & 0.378 & 55.8 & -- & -- \\
Easi3R~\cite{chen2025easi3r}            & \textcolor{red}{\ding{55}}  & 0.110 & 0.042 & 0.758 & 0.377 & 55.9 & -- & -- \\
VGGT~\cite{wang2025vggt}                & \textcolor{red}{\ding{55}}  & 0.172 & 0.062 & 0.471 & 0.299 & 63.8 & -- & -- \\
$\pi^3$~\cite{wang2026pi}               & \textcolor{red}{\ding{55}}  & \textbf{0.073} & \textbf{0.038} & \textbf{0.288} & \textbf{0.233} & \textbf{66.4} & -- & -- \\
\midrule
Spann3R~\cite{wang20253d}               & \textcolor{teal}{\ding{51}} & 0.329 & 0.110 & 4.471 & 0.622 &42.6 & -- & -- \\
Point3R~\cite{wu2025point3r}            & \textcolor{teal}{\ding{51}} & 0.351 & 0.128 & 1.822 & 0.452 & 48.9 & 0.78 & 17.1 \\
StreamVGGT~\cite{zhuo2025streaming}     & \textcolor{teal}{\ding{51}} & 0.251 & 0.149 & 1.894 & \textbf{0.323} & \textbf{65.7} & -- & -- \\
STream3R~\cite{lan2026streamr}         & \textcolor{teal}{\ding{51}} & 0.213 & 0.076 & 0.868 & 0.478 & 51.1 & 1.04 & 21.0 \\
CUT3R~\cite{wang2025continuous}         & \textcolor{teal}{\ding{51}} & 0.209 & \textbf{0.069} & \textbf{0.624} & 0.433 & 46.9 & 1.03 & 23.6 \\
TTT3R~\cite{chen2026tttr}               & \textcolor{teal}{\ding{51}} & 0.210 & 0.091 & 0.720 & 0.405 & 48.9  & 0.98 & 23.2 \\
TTSA3R~\cite{zheng2026ttsa3r}  & \textcolor{teal}{\ding{51}}&0.210 & 0.085 & 0.765 & 0.402 & 49.8 & \textbf{0.96} & 24.6\\
\ours                      & \textcolor{teal}{\ding{51}} & \textbf{0.180} & 0.074 & 0.860 & 0.438 & 44.1 & 1.10 & 24.6 \\
\ours + TTT3R  & \textcolor{teal}{\ding{51}} & 0.20 & 0.091 & 0.720 & 0.414 & 46.7 & 1.02 & 25.2\\
\ours + TTSA3R & \textcolor{teal}{\ding{51}}& 0.20 & 0.087 & 0.754 & 0.413 & 47.0 & 0.99 & \textbf{25.8} \\
\bottomrule
\end{tabular}
}
\label{tab:short_camera_pose_depth_sintel}
\end{table*}

\begin{table}[htbp]
\centering
\caption{\textbf{Video Depth Estimation on \textbf{short} sequence.} We evaluate scale-invariant depth accuracy on KITTI and Bonn datasets. Methods that require global alignment are denoted as ``GA''.}
\resizebox{\linewidth}{!}{
\begin{tabular}{l c cc cc}
\toprule
\textbf{Method} & \textbf{Online}
& \multicolumn{2}{c}{\textbf{KITTI (110 frames)}}
& \multicolumn{2}{c}{\textbf{Bonn (110 frames)}} \\
\cmidrule(lr){3-4} \cmidrule(lr){5-6}
& & Abs Rel $\downarrow$ & $\delta_{1.25} \uparrow$
& Abs Rel $\downarrow$ & $\delta_{1.25}\uparrow$ \\
\midrule

 DUSt3R-GA~\cite{DUSt3R}             & \textcolor{red}{\ding{55}}  & 0.144 & 81.3 & 0.155 & 83.3 \\
MASt3R-GA~\cite{leroy2024grounding} & \textcolor{red}{\ding{55}}  & 0.183 & 74.5 & 0.252 & 70.1 \\
MonST3R-GA~\cite{zhang2025monstr}   & \textcolor{red}{\ding{55}}  & 0.168 & 74.4 & 0.067 & 96.3 \\
Easi3R~\cite{chen2025easi3r}        & \textcolor{red}{\ding{55}}  & 0.102 & 91.2 & 0.059 & 97.0 \\
VGGT~\cite{wang2025vggt}            & \textcolor{red}{\ding{55}}  & \textbf{0.070} & \textbf{96.5} & \textbf{0.055} & \textbf{97.1} \\
\cmidrule(lr){1-6}
Spann3R~\cite{wang20253d}           & \textcolor{teal}{\ding{51}} & 0.198 & 73.7 & 0.144 & 81.3 \\
Point3R~\cite{wu2025point3r}        & \textcolor{teal}{\ding{51}} & 0.136 & 84.2 & 0.060 & 96.0 \\
STREAM3R$^\alpha$~\cite{lan2026streamr} & \textcolor{teal}{\ding{51}} & 0.116 & 89.6 & 0.075 & 94.1 \\
StreamVGGT~\cite{zhuo2025streaming} & \textcolor{teal}{\ding{51}} & 0.173 & 72.1 & \textbf{0.059} & \textbf{97.2} \\
CUT3R~\cite{wang2025continuous}  & \textcolor{teal}{\ding{51}} &  0.122 & 87.5 & 0.076 & 94.0\\
TTT3R~\cite{chen2026tttr}           & \textcolor{teal}{\ding{51}} & 0.114 & 90.4 & 0.069 & 95.5\\
TTSA3R~\cite{zheng2026ttsa3r}       & \textcolor{teal}{\ding{51}} & 0.110 & \textbf{91.0} & 0.064 & 96.4\\
\ours & \textcolor{teal}{\ding{51}} &  0.113 & 89.4 & 0.074 & 94.3 \\
\ours + TTT3R         & \textcolor{teal}{\ding{51}} & 0.110 & 90.3 & 0.067 & 95.6\\
\ours + TTSA3R          & \textcolor{teal}{\ding{51}} & \textbf{0.109} & \textbf{91.0} & 0.065 & 95.9\\
\bottomrule

\end{tabular}
}
\label{tab:video_depth_kitti_bonn_short}
\end{table}

\paragraph{3D Reconstruction.}
\label{app:3d_recon}

As shown in \cref{tab:reconstruction_merged}, \modelname yields improvements on the NRGBD dataset that are comparable to those observed on the 7-Scenes dataset.

\begin{table}[htbp]
\centering
\caption{Quantitative evaluation of multi-view reconstruction. OOM denotes out-of-memory. \textbf{Bold} denotes the best result in each column within each dataset, and \colorbox{lightgreen}{green} indicates that our model matches or outperforms its corresponding base model. For the NRGBD dataset, one test sequence contains fewer than 400 frames.}
\label{tab:reconstruction_merged}
\resizebox{0.5\textwidth}{!}{
\begin{tabular}{l cc cc cc cc cc}
\toprule
\multirow{2}{*}{\textbf{Method}} & \multicolumn{2}{c}{\textbf{200 frames}} & \multicolumn{2}{c}{\textbf{250 frames}} & \multicolumn{2}{c}{\textbf{300 frames}} & \multicolumn{2}{c}{\textbf{350 frames}} & \multicolumn{2}{c}{\textbf{400 frames}} \\
\cmidrule(lr){2-3} \cmidrule(lr){4-5} \cmidrule(lr){6-7} \cmidrule(lr){8-9} \cmidrule(lr){10-11}
 & CD $\downarrow$ & NC $\uparrow$ & CD $\downarrow$ & NC $\uparrow$ & CD $\downarrow$ & NC $\uparrow$ & CD $\downarrow$ & NC $\uparrow$ & CD $\downarrow$ & NC $\uparrow$ \\
\midrule
\multicolumn{11}{c}{\textbf{NRGBD Dataset}} \\
\midrule
VGGT (offline) & OOM & OOM & OOM& OOM& OOM& OOM& OOM& OOM& OOM& OOM \\
Stream-VGGT & OOM & OOM & OOM& OOM& OOM& OOM& OOM& OOM& OOM& OOM \\
\midrule
CUT3R & 0.081 & 0.602 & 0.130 & 0.594 & 0.162 & 0.576 & 0.188 & 0.566 & 0.220 & 0.552 \\
\ours &  \gb  0.060 &  \gb  0.612 &  \gb  0.085 &  \gb  0.609 &  \gb  0.101 &  \gb  0.592 &  \gb  0.114 &  \gb  0.585 &  \gb  0.135 &  \gb  0.577 \\
\midrule
TTT3R & 0.037 & 0.626 & 0.049 & 0.621 & 0.065 & 0.605 & 0.085 & 0.599 & 0.104 & 0.595 \\
\ours + TTT3R &  \gb  0.037 &  0.625 &  \gb  0.048 &  \gb  0.621 &  \gb  0.062 &  \gb  0.612 &  \gb  0.062 &  \gb  0.608 &  \gb  0.064 &  \gb  0.608 \\
\midrule
TTSA3R & \textbf{0.031} & \textbf{0.630} & \textbf{0.042} & \textbf{0.622} & 0.057 & \textbf{0.616} & 0.060 & 0.611 & 0.069 & 0.609 \\
\ours + TTSA3R &   0.032 &   0.625 &  \gb  \textbf{0.042} &   0.621 &  \gb  \textbf{0.052} &  \gb  \textbf{0.616} &  \gb  \textbf{0.057} &  \gb  \textbf{0.614} &  \gb  \textbf{0.061} &  \gb  \textbf{0.612} \\
\bottomrule
\end{tabular}}
\end{table}

\subsection{Training Datasets}
\label{app:train_data}
Following the final stage of training for CUT3R, we only use multi-view datasets during training. Our fine-tuning follows the same training data configuration as CUT3R, using the dataset mixture defined in its \texttt{dpt\_512\_vary\_4\_64.yaml} config.
Specifically, the training set consists of Co3Dv2~\cite{reizenstein2021common}, WildRGBD~\cite{xia2024rgbd}, ARKitScenes~\cite{baruch2021arkitscenes}, ARKitScenes-HighRes~\cite{baruch2021arkitscenes}, ScanNet++~\cite{scannetpp-YeshwanthLND23}, ScanNet~\cite{scannet-DaiCSHFN17}, HyperSim~\cite{roberts2021hypersim}, BlendedMVS~\cite{yao2020blendedmvs}, MegaDepth~\cite{li2018megadepth}, MapFree~\cite{arnold2022map}, Waymo~\cite{sun2020scalability}, VirtualKITTI2~\cite{cabon2020virtual}, Unreal4K~\cite{tosi2021smd}, TartanAir~\cite{wang2020tartanair}, DL3DV~\cite{ling2024dl3dv}, Cop3D~\cite{sinha2023common}, MVImgNet~\cite{yu2023mvimgnet}, RealEstate10K~\cite{zhou2018stereo}, OmniObject3D~\cite{wu2023omniobject3d}, Dynamic Replica~\cite{karaev2023dynamicstereo}, Spring~\cite{mehl2023spring}, BEDLAM~\cite{black2023bedlam}, MVS-Synth~\cite{huang2018deepmvs}, PointOdyssey~\cite{zheng2023pointodyssey}, UASOL~\cite{bauer2019uasol}, and Matterport3D~\cite{chang2017matterport3d}.

\subsection{Social Impacts}
\label{app:social}

\paragraph{Positive Social Impact.}
Our real-time streaming 3D reconstruction framework can broaden the applicability of online 3D perception in real-world scenarios. By maintaining a fixed-size memory while improving long-sequence temporal consistency, our model is well suited for applications such as robotics, AR/VR, and embodied agents, where systems need to process continuous visual streams efficiently and reliably. Its improved stability over long sequences may further benefit navigation, scene understanding, and interactive 3D mapping in dynamic real-world environments.

\paragraph{Potential Negative Impact.}
We do not foresee significant misuse risks specific to our method beyond the general considerations associated with 3D reconstruction and perception technologies. As with other 3D perception systems, deployment in real-world environments may raise privacy concerns if visual streams are collected or processed without user consent, especially in private indoor spaces. In addition, although our model improves long-sequence performance, occasional failure cases may still occur under challenging conditions such as severe motion blur, rapid viewpoint changes, textureless regions, or highly dynamic scenes. Practical deployments should therefore follow standard privacy-preserving data collection practices and include downstream validation mechanisms to detect unreliable predictions.

{
    \small
    \bibliographystyle{ieeenat_fullname}
    \bibliography{main}
}

\end{document}